\documentclass{article}

\usepackage{microtype}
\usepackage{graphicx}
\usepackage{subfigure}
\usepackage{booktabs} 
\usepackage{amsmath}
\usepackage{amssymb}
\usepackage{url}
\usepackage{enumerate}

\usepackage{hyperref}
\usepackage{breqn}
\usepackage{dirtytalk}

\usepackage[accepted]{icml2023}

\icmltitlerunning{Dirichlet Diffusion Score Model for Biological Sequence Generation}

\begin{document}

\twocolumn[
\icmltitle{Dirichlet Diffusion Score Model for Biological Sequence Generation}



\icmlsetsymbol{equal}{*}

\begin{icmlauthorlist}
\icmlauthor{Pavel Avdeyev}{utsw}
\icmlauthor{Chenlai Shi}{utsw}
\icmlauthor{Yuhao Tan}{utsw}
\icmlauthor{Kseniia Dudnyk}{utsw}
\icmlauthor{Jian Zhou}{utsw}
\end{icmlauthorlist}

\icmlaffiliation{utsw}{Lyda Hill Department of Bioinformatics, University of Texas Southwestern Medical Center, USA}

\icmlcorrespondingauthor{Jian Zhou}{jian.zhou@utsouthwestern.edu}

\icmlkeywords{Machine Learning, ICML}

\vskip 0.3in
]



\printAffiliationsAndNotice{}  

\begin{abstract}
Designing biological sequences is an important challenge that requires satisfying complex constraints and thus is a natural problem to address with deep generative modeling. Diffusion generative models have achieved considerable success in many applications. Score-based generative stochastic differential equations (SDE) model is a continuous-time diffusion model framework that enjoys many benefits, but the originally proposed SDEs are not naturally designed for modeling discrete data. To develop generative SDE models for discrete data such as biological sequences, here we introduce a diffusion process defined in the probability simplex space with stationary distribution being the Dirichlet distribution. This makes diffusion in continuous space natural for modeling discrete data. We refer to this approach as Dirchlet diffusion score model. We demonstrate that this technique can generate samples that satisfy hard constraints using a Sudoku generation task. This generative model can also solve Sudoku, including hard puzzles, without additional training. Finally, we applied this approach to develop the first human promoter DNA sequence design model and showed that designed sequences share similar properties with natural promoter sequences.

\end{abstract}

\section{Introduction}
\label{submission}

Diffusion probabilistic models are a family of models that reverse diffusion process to generate data from noise. Score-based generative stochastic differential equation (SDE) is a type of continuous-time diffusion model that has many desirable properties, such as allowing likelihood evaluation through a connection to a probability flow ordinary differential equation (ODE) and flexibility in sampling approaches. However, the originally proposed generative SDEs are not directly suitable for modeling discrete data. Recent works have proposed methods for adapting diffusion models to discrete data \cite{campbell2022continuous,chen2022analog,sun2022score,austin2021structured,hoogeboom2021argmax,hoogeboom2021autoregressive}, including continuous-time diffusion in discrete space \cite{campbell2022continuous}, but no methods are formulated within the continuous-time SDE diffusion framework \cite{song2020score}, except for quantization-based methods. In this manuscript, we propose a general mechanism to extend this approach to discrete data, while allowing continuous-time diffusion in probability simplex space. Specifically, designed to utilize the natural connection between Dirichlet distribution and discrete data, we consider continuous-time diffusion within the probability simplex for which the stationary distribution is Dirichlet distribution. Forward diffusion (data-to-noise) of discrete data starts from the vertices of the probability simplex space, and diffuses continuously in the same space, and the continuous-discrete space gap can be bridged with a latent variable interpretation.

While our intended application is in biological sequence generation, we evaluated our, Dirichlet diffusion score model (DDSM)\footnote{Code available at https://github.com/jzhoulab/ddsm}, on a range of discrete data generation tasks to better understand its performance. In addition to demonstrating competitive performance on a small benchmark dataset, binarized MNIST, we applied it to generating Sudoku puzzles to test for its ability in generating highly structured data with strong constraints. The model can not only generate but also solve Sudoku puzzles including hard puzzles, which is the first time this is achieved with a purely generative modeling approach. Finally, we applied DDSM to a real-world application in biological sequence generation. Specifically, we developed the first model for designing human promoter DNA sequences that drive gene expression, and demonstrate that it designs diverse sequences comparable to human genome promoter sequences.

\section{Background}
\subsection{Score-based Generative Modeling with SDE}\label{section_scoresde}

It\^{o} diffusion process is defined as
$$\mathrm{d} \mathbf{x} = \mathbf{f}(\mathbf{x}, t)\mathrm{d}t +\mathbf{G}(\mathbf{x}, t)\mathrm{d}\mathbf{w},$$
where $\mathbf{w}$ is the standard Wiener process (a.k.a., Brownian motion), the drift coefficient $f(\cdot,t): \mathbb{R}^n \to \mathbb{R}^n$ and diffusion coefficient $G(\cdot, t): \mathbb{R}^n \to \mathbb{R}^n$ are vector-valued functions of $x^t$. \citealt{song2020score} exploited the remarkable result by \citealt{anderson1982reverse} that the time-reversal of this diffusion process can be obtained by the following SDE:
\begin{equation}\label{eq:reverse_sde}
\begin{aligned}
    \mathrm{d}\mathbf{x}=\biggl\{&\mathbf{f}(\mathbf{x}, t)-\nabla \cdot\left[\mathbf{G}(\mathbf{x}, t) \mathbf{G}(\mathbf{x}, t)^{\top}\right] \\
    &-\mathbf{G}(\mathbf{x}, t) \mathbf{G}(\mathbf{x}, t)^{\top} \nabla_{\mathbf{x}} \log p_{t}(\mathbf{x})\biggr\} \mathrm{d}t \\
    &+\mathbf{G}(\mathbf{x}, t) \mathrm{d} \overline{\mathbf{w}},
\end{aligned}
\end{equation}
where $\nabla \cdot\left[\mathbf{G}(\mathbf{x}, t) \mathbf{G}(\mathbf{x}, t)^{\top}\right]$ indicates row-sums of element-wise derivative with respect to $x$. The corresponding probability flow ODE is defined as
\begin{equation}\label{eq:probability_flow_ode}
\begin{aligned}
\mathrm{d} \mathbf{x}=\biggl\{&\mathbf{f}(\mathbf{x}, t)-\frac{1}{2} \nabla \cdot\left[\mathbf{G}(\mathbf{x}, t) \mathbf{G}(\mathbf{x}, t)^{\top}\right]\\
&-\frac{1}{2} \mathbf{G}(\mathbf{x}, t) \mathbf{G}(\mathbf{x}, t)^{\top} \nabla_{\mathbf{x}} \log p_{t}(\mathbf{x})\biggr\} \mathrm{d} t.
\end{aligned}
\end{equation}
It gives the same distribution at time $t$ as the reverse-time SDE. Both the reverse-time SDE and the probability flow ODE can be sampled from given $\nabla_{\mathbf{x}} \log p_{t}(\mathbf{x})$ or the score of $p_{t}(\mathbf{x})$. Therefore, learning the reverse diffusion becomes the problem of learning the score function, which is usually parameterized as a neural network known as the \emph{score model}. The training loss is the score matching loss 
\begin{equation}\label{score_matching_loss}
\hspace{-0.8em}\int_{0}^{T} \mathbb{E}_{p_t(\mathbf{x^t}) }\left[\lambda(t)\left\|\nabla_{\mathbf{x}} \log p\left(\mathbf{x^t}\right)-s_{\boldsymbol{\theta}}\left(\mathbf{x^t}, t\right)\right\|_{2}^{2}\right] \mathrm{d}t,
\end{equation}
where $s_{\boldsymbol{\theta}}\left(\mathbf{x}, t\right)$ is the score model to be trained, and $\lambda(t)$ is a positive weighting function. The loss is equivalent to the denoising score matching loss 
\begin{equation}\label{eq:denoising_score_matching_loss}
\begin{aligned}
\int_{0}^{T} \mathbb{E}_{p_0(\mathbf{x^0}) p\left(\mathbf{x^t} \mid \mathbf{x^0}\right)}\biggl[\lambda(t) \Bigl\|\nabla_{\mathbf{x}} &\log p\left(\mathbf{x^t} \mid \mathbf{x^0}\right)\\
&-s_{\boldsymbol{\theta}}\left(\mathbf{x^t}, t\right)\Bigr\|_{2}^{2}\biggr] \mathrm{d}t,
\end{aligned}
\end{equation}
which is used in practice because $\nabla_{\mathbf{x}} \log p\left(\mathbf{x^t} \mid \mathbf{x^0}\right)$ is usually easier to compute. Forward diffusion processes considered so far have Gaussian stationary distribution and are applicable for continuous data in $\mathbb{R}^n$.

\subsection{Univariate Jacobi Diffusion Process}\label{section:jacobi_diff_proc}
We consider Jacobi diffusion process\footnote{It is also known as the univariate Wright-Fisher diffusion} in the following form 
\begin{equation}\label{eq:jacobi_diffusion}
\mathrm{d}\mathbf{x} = \frac{s}{2}[a(1-\mathbf{x})-b\mathbf{x}]\mathrm{d}t + \sqrt{s\mathbf{x}(1-\mathbf{x})} \mathrm{d}\mathbf{w},
\end{equation}
where $0\leq x\leq 1$, $s > 0$ is the speed factor, and $a > 0$, $b > 0$ determines the stationary distribution $\textbf{Beta}(a, b)$. We usually use $s=1$ or $s=\frac{2}{a+b}$ (see Appendix \ref{a:speed_factor} for discussion of choices). Note that when $x$ approaches $0$ or $1$, the diffusion coefficient converges to $0$ and the drift coefficient converges to $a$ or $-b$, keeping the diffusion within $[0,1]$.

The spectral expansion of the transition density function was derived by \citealt{kimura1955stochastic, Kimura1957-gm}. Hence, the diffused density at any time $t$ is computed by the following formula:
\begin{equation} \label{eq:transition_density_function}
\begin{aligned}
\hspace{-0.6em}&~p_{a,b}(x^t|x^0)\\
\hspace{-1em}&= \mathcal{B}_{a,b}\left(x^t\right) \sum_{n=0}^{\infty}\frac{e^{\lambda_n t}}{d_n} R_n^{(a,b)}\left(x^0\right)R_n^{(a,b)}\left(x^t\right) \\
\hspace{-1em}&= \mathcal{B}_{a,b}\left(x^t\right) \left(1+\sum_{n=1}^{\infty}\frac{e^{\lambda_n t}}{d_n} R_n^{(a,b)}\left(x^0\right)R_n^{(a,b)}\left(x^t\right)\right),
\end{aligned}
\end{equation}
where $\mathcal{B}_{a,b}(x_t)$ is the $\textbf{Beta}(a,b)$ density, $R_n^{(a, b)}(x)$ denotes the $n$-th order modified Jacobi polynomial of order $n$ and are eigenfunctions of the generator of the Jacobi diffusion process \cite{steinrucken2013explicit,griffiths2010diffusion}. The corresponding eigenvalues are $\lambda_n=-\frac{1}{2}sn(n-1+a+b)$. 
The gradient of the log transition density function can be computed via automatic differentiation.

\section{Diffusion Processes for Generative SDE Modeling of Discrete Data}

\begin{figure*}[ht]
\begin{center}
\centerline{\includegraphics[width=0.85\textwidth]{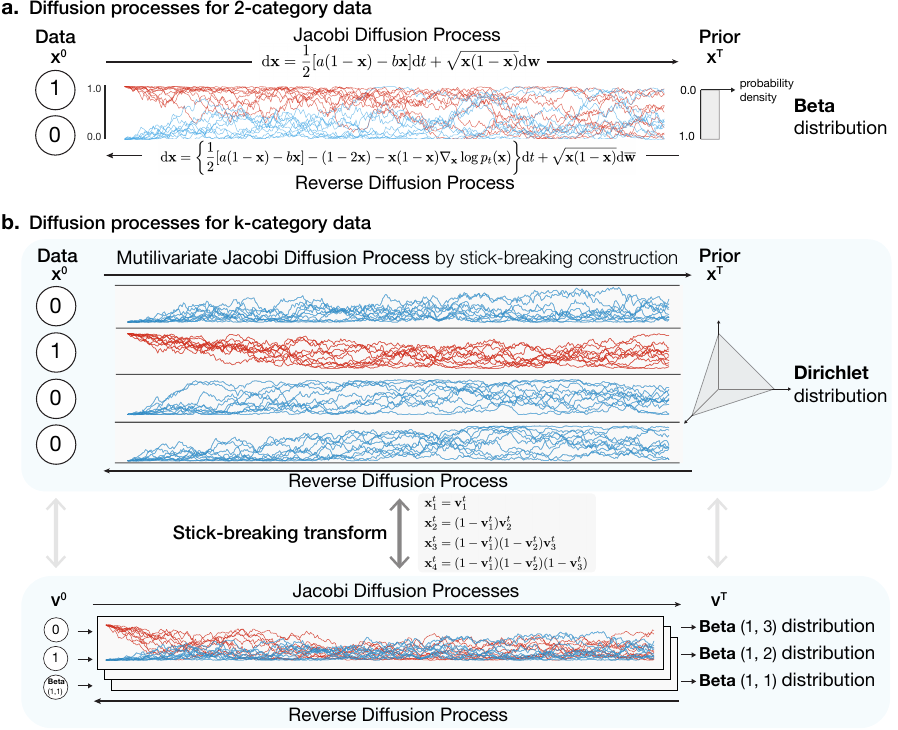}}
\vskip -0.15in
\caption{Schematic overview of forward and reverse diffusion SDEs for Dirichlet diffusion score model. Forward and reverse diffusion SDEs for 2-category data (a) and k-category data (b) by stick-breaking construction are shown.}
\label{fig:diagram}
\end{center}
\vskip -0.35in
\end{figure*}

\subsection{Forward Diffusion SDE for Two-Category Data}
Using the univariate Jacobi diffusion as the forward diffusion process provides a natural generalization of the score-based generative SDE approach (Section~\ref{section_scoresde}) to discrete data with two categories, encoded as $0$ and $1$. The forward diffusion starting from $0$ or $1$ at the initial timepoint will continuously diffuse in the $[0,1]$ interval and converge to a Beta stationary distribution (see Fig.~\ref{fig:diagram}a). If $a=1$ and $b=1$ in Eq. \ref{eq:jacobi_diffusion} and \ref{eq:transition_density_function} then the Beta stationary distribution is \textbf{Beta}(1,1) (i.e., a uniform distribution in the interval $[0,1]$). 

The score-based generative SDE model can be trained via the denoising score matching objective (see Eq.~\ref{eq:denoising_score_matching_loss}) following the transition density formula (see Eq.~\ref{eq:transition_density_function}). By combining equations~\ref{eq:reverse_sde} and~\ref{eq:jacobi_diffusion}, we have the following reverse-time SDE: 
\begin{equation}\label{eq:reverse_sde_jacobi}
\begin{aligned}
    \mathrm{d}\mathbf{x}=\biggl\{&\frac{s}{2}[a(1-\mathbf{x})-b\mathbf{x}] -
    s(1-2\mathbf{x}) \\ &-s\mathbf{x}(1-\mathbf{x}) \nabla_{\mathbf{x}} \log p_{t}(\mathbf{x})\biggr\} \mathrm{d}t \\ &+\sqrt{s\mathbf{x}(1-\mathbf{x})} \mathrm{d} \overline{\mathbf{w}}.
\end{aligned}
\end{equation}
Replacing $\nabla_{\mathbf{x}} \log p_{t}(\mathbf{x})$ with score model $s_{\boldsymbol{\theta}}\left(\mathbf{x^t}, t\right)$ allows sampling from the trained model via reverse diffusion.

\subsection{Forward Diffusion SDE for $k$-Category Data}

To model discrete variables with $k$ categories, e.g. DNA sequence (with four bases A,C,G,T) or protein sequence ($20$ amino acid residues), we need to consider diffusion in probability simplex. 

We seek to use a multivariate diffusion process over the probability simplex for which the stationary distribution is Dirichlet distribution (see Fig.~\ref{fig:diagram}b). Jacobi diffusion process converges to Beta stationary distribution, a univariate special case of Dirichlet distribution. Using the connection between Beta distribution and Dirichlet distribution, we construct a multivariate diffusion process on probability simplex that converges to Dirichlet distribution with $k-1$ independent univariate Jacobi diffusion processes by a classical stick-breaking construction
$$\mathbf{x}_1^t=\mathbf{v}_1^t,\, \mathbf{x}_2^t=(1-\mathbf{v}_1^t)\mathbf{v}_2^t, \, \mathbf{x}_3^t=(1-\mathbf{v}_1^t)(1-\mathbf{v}_2^t)\mathbf{v}_3^t, \, \cdots,$$
where $\mathbf{v}_1^t, \mathbf{v}_2^t, \dots, \mathbf{v}_{k-1}^t$ are drawn from independent Jacobi diffusion processes at time $t$. Thus, we obtain a multivariate diffusion process with Dirichlet stationary distribution using $\mathbf{x}_1^t, \mathbf{x}_2^t, \dots, \mathbf{x}_{k}^t$.

For notation simplicity, we will use $\mathbf{v}$ and $\mathbf{x}$ to indicate the $k-1$ and $k$ dimensional representations for the rest of the manuscript, respectively. The conversion between $\mathbf{v}$ and $\mathbf{x}$ is done by stick-breaking transform and its inverse. 

For obtaining any Dirichlet stationary distribution, we parameterize the Jacobi diffusion process. For example, for the stationary distribution to be the flat Dirichlet distribution $\textbf{Dir}(1,1,\dots,1)$ (i.e., the uniform distribution over the probability simplex), we need to choose the Jacbobi diffusion processes with stationary distributions $\textbf{Beta}(1, k-1), \textbf{Beta}(1, k-2),\dots, \textbf{Beta}(1, 1)$ for $\mathbf{v}_1, \mathbf{v}_2, \dots, \mathbf{v}_{k-1}$ of stick-breaking construction. This can be simply achieved by choosing parameters $a,b$ in equation~\ref{eq:jacobi_diffusion} to be $(1, k-1), (1, k-2),\dots, (1, 1)$. 

We refer to this multivariate diffusion process as \emph{multivariate Jacobi diffusion process by stick-breaking construction}, and the generative modeling approach with this diffusion process as \emph{Dirichlet diffusion score model}. An infinite dimensional form of this diffusion process with a more general distribution family has been proposed as the GEM process \cite{feng_wang_2007}. The proposed process is a finite-dimensional version of the GEM process. We note that other forms of diffusion processes for which the stationary distribution is Dirichlet distribution exist (see \citealt{Steinrucken2013-sp, bakosi2013}). However, they are much more computationally expensive to use as forward diffusion processes since they cannot be decomposed into independent univariate diffusion processes.

\subsection{Score Matching Training for $k$-Category Discrete Data}\label{section:score_matching_training_K}
Using the multivariate Jacobi diffusion by stick-breaking construction as the forward diffusion process allows us to train score-based diffusion model for $k$-category discrete data.

The initial value of the diffusion in $\mathbf{x}$ space is set to be the discrete data represented by $k$-dimensional one-hot encoding such as $(0,0,1,\dots,0)$. To sample from the forward diffusion, we first map the initial values of $\mathbf{x}$ to $\mathbf{v}$ space via inverse stick-breaking transform. For all dimensions of $\mathbf{v}$ after the first $1$ that is undetermined by inverse stick-breaking transform given $\mathbf{x}$, we consider them as drawn from corresponding stationary Beta distribution. Explicitly drawing the Beta samples for initial values are not needed since the density remains stationary and the samples and scores at any time are directly computed from the Beta distributions. 

The forward diffusion samples in $\mathbf{v}$ space at any time $t$ are drawn from the Jacobi diffusion processes (for dimension with deterministic initial values) and stationary Beta distributions (for dimension with undetermined initial values). The scores for the transition density function in denoising score-matching loss (Eq. \ref{eq:denoising_score_matching_loss}) are then computed from the corresponding Jacobi diffusion transition density function and Beta density function. 

By applying the change-of-variable conversion, we can equivalently perform score matching in either $\mathbf{v}$ space or $\mathbf{x}$ space since we can convert between the score of $\mathbf{x}$ to the score of $\mathbf{v}$: 
\begin{equation*} 
\begin{split}
\frac{\partial{\log p_{\mathbf{x}}(\mathbf{x})}}{\partial \mathbf{x}} &=
\left(\frac{ \partial{\log p_{\mathbf{v}}(\mathbf{v})}}{\partial \mathbf{v}} + \frac{\partial \log |\det{\frac{\partial \mathbf{v}}{\partial \mathbf{x}}}|}{\partial \mathbf{v}} \right)\frac{\partial \mathbf{v}}{\partial \mathbf{x}}, 
\end{split}
\end{equation*}
\begin{equation*} 
\begin{split}
 &\frac{\partial{\log p_{\mathbf{v}}(\mathbf{v})}}{\partial \mathbf{v}} = 
\frac{\partial{\log p_{\mathbf{x}}(\mathbf{x})}}{\partial \mathbf{x}} \frac{\partial \mathbf{x}}{\partial \mathbf{v}}-\frac{\partial \log |\det \frac{\partial \mathbf{v}}{\partial \mathbf{x}}|}{\partial \mathbf{v}}.
\end{split}
\end{equation*}

The score model $s_{\boldsymbol{\theta}}\left(\mathbf{x^t}, t\right)$ is more naturally formulated as a function of $\mathbf{x}$, and we choose to compute score-matching loss in $\mathbf{v}$ space because of the diagonal form of diffusion coefficient.

Once the score model is learned, sampling from the reverse diffusion process in $\mathbf{v}$ space is nearly identical to sampling from multiple univariate reverse diffusion processes as in Equation \ref{eq:reverse_sde_jacobi}, except for that the score model takes all dimensions of $\mathbf{v}$ as input (after converting to $\mathbf{x}$ space).

\subsection{Weighting Function for Score Matching Loss of General SDE}
The choice of weighting function $\lambda(t)$ in score matching loss (Eq. \ref{score_matching_loss} and \ref{eq:denoising_score_matching_loss}) has previously been studied \cite{Song2021-zl,Huang2021-jt}. With the assumption that the scalar diffusion coefficient $g(\mathbf{v}, t)$ of the forward diffusion process does not depend on the value $\mathbf{v}$ being diffused, $\lambda(t)=g(t)^2$ is shown to be the likelihood weighting \cite{Song2021-zl}. Minimizing the loss function with this weighting is equivalent to maximizing the ELBO \cite{Huang2021-jt}. However, this assumption about $g(\mathbf{v}, t)$ does not hold for the Jacobi diffusion process.

Here we motivate the use of 
\begin{equation} 
\begin{split} \label{eq:weightedloss}
L(\mathbf{v}, t) =& \left\| \frac{\partial{\log p_{\mathbf{v}}(\mathbf{v})}}{\partial \mathbf{v}} - \frac{\partial{\log q_{\mathbf{v}}(\mathbf{v})}}{\partial \mathbf{v}} \right\|_{\mathbf{G}\mathbf{G}^T}^2 \\ 
=& 
\left( \frac{\partial{\log p_{\mathbf{v}}(\mathbf{v})}}{\partial \mathbf{v}} - \frac{\partial{\log q_{\mathbf{v}}(\mathbf{v})}}{\partial \mathbf{v}} \right) \mathbf{G}(\mathbf{v}, t) \\
&\mathbf{G}(\mathbf{v}, t)^T \left( \frac{\partial{\log p_{\mathbf{v}}(\mathbf{v})}}{\partial \mathbf{v}} - \frac{\partial{\log q_{\mathbf{v}}(\mathbf{v})}}{\partial \mathbf{v}} \right)^T
\end{split}
\end{equation}
to be the general form of weighted score-matching loss for any SDE with matrix-form diffusion coefficient $\mathbf{G}(\mathbf{v},t)$, from the argument that the loss function should satisfy the property of invariance under change-of-variable, which is satisfied by likelihood function. In Appendix \ref{a:weighting_function}, we show that this loss function is invariant to change-of-variable by any bijective, differentiable function $\mathbf{x}=h(\mathbf{v})$, while the unweighted loss is not.
If $\mathbf{G}$ is scalar or diagonal and does not depend on $\mathbf{v}$, we recover the likelihood weighting from \citealt{song2020score}.

\subsection{Likelihood Computation}
After the model is trained with score-matching, we can estimate likelihood using probability flow ODE (see Eq.~\ref{eq:probability_flow_ode}). Our formulation allows both computing the likelihood from the continuous distribution over the probability simplex, and computing a variation lower-bound of the likelihood of discrete data.

\subsubsection*{Likelihood Computation for Continuous Variable in Probability Simplex}

We will first estimate the likelihood in the $\mathbf{v}$ space, which can be easily converted to likelihood in the $\mathbf{x}$ space using the probability change-of-variable formula~\cite{Chen2018-pc}. 

Following \citealt{Song2021-zl}, the probability flow ODE for a Jacobi diffusion process is:
\begin{equation*}
\begin{aligned}
d \mathbf{v} =& \biggl\{ \frac{s}{2}[a(1-\mathbf{v})-b\mathbf{v}] \\
&- \frac{s}{2} (1-2\mathbf{v}) - \frac{s}{2} \mathbf{v}(1-\mathbf{v}) \mathbf{s}_\theta(\mathbf{v},t) \biggr\} d t = \tilde{f}dt
\end{aligned}
\end{equation*}

By the instantaneous change-of-variable formula, we have:
\begin{equation*}
p_0 (\mathbf{v^0}) = e^{\int_0^t \operatorname{tr}\left(\nabla_{\mathbf{v}}{ \tilde{\mathbf{f}}(v^t)} \right) d t} p_t(\mathbf{v^t}).
\end{equation*}

The trace of Jacobian can be unbiasedly approximated by Hutchinson's estimator
\begin{equation*}
\operatorname{tr}\left(\nabla_{\mathbf{v}}{ \tilde{\mathbf{f}}(v^t)}\right) = \mathbf{E}_{\boldsymbol\epsilon \sim \mathcal{N}(\mathbf{0}, \mathbf{I})}[\boldsymbol\epsilon^T \nabla_{\mathbf{v}}{ \tilde{\mathbf{f}}(v^t)} \boldsymbol \epsilon]
\end{equation*} 
and $p_t(\mathbf{v^t})$ is computed with the stationary distribution. $p_0(\mathbf{v^0})$ is converted to $p_0(\mathbf{x^0})$ by applying the change-of-variable formula to obtain the likelihood.

Since this continuous-space likelihood is not directly comparable with discrete-space likelihood, next we will derive an evidence lower bound (ELBO) that allows direct comparison with likelihood in discrete data space.

\subsubsection*{Bounding Discrete Data Likelihood with Variational Lowerbound}\label{section: discretelikelihood}

To obtain a variational lowerbound for discrete likelihood, we consider the continuous variable $\mathbf{x}$ in probability simplex space, drawn from the reverse diffusion process, as directly parameterizing categorical distributions. The discrete data $\mathbf{y}$ are drawn from these categorical distributions. Thus we obtain the discrete likelihood by marginalizing over $\mathbf{x}$ $p(\mathbf{y})=\int{p^{\text{Cat}}(\mathbf{y}|\mathbf{x})p(\mathbf{x})d\mathbf{x}}$. 

While this is generally intractable computationally, we use the variational lowerbound ELBO
\begin{equation*}
\begin{aligned}
\log p\left(\mathbf{y}\right) \geq \mathbb{E}_{q^{\text{Diff}}(\mathbf{\mathbf{x}} | \mathbf{y})}\bigl[-\log q^{\text{Diff}}(\mathbf{x} | \mathbf{y})&+\log p^{\text{Cat}}(\mathbf{y} | \mathbf{x}) \\
&+\log p^{\text{ODE}}(\mathbf{x})\bigr],
\end{aligned}
\end{equation*}
where $q^{\text{Diff}}(\mathbf{x}|\mathbf{y})$ is the density of forward diffusion from $\mathbf{y}$ at time $t_{\tilde{0}}$, with $t_{\tilde{0}}$ chosen to be close to 0. 
$p^{\text{Cat}}(\mathbf{y}|\mathbf{x})$ is the categorical distribution likelihood. $p^{\text{ODE}}(\mathbf{x})$ is the continuous-space likelihood of probability flow ODE as described in the previous subsection, but with the lower end of time being $t_{\tilde{0}}$ instead of 0. This expectation is unbiasedly estimated by sampling from the forward diffusion process. This ELBO formulation is chosen so that the diffusion model training will also minimize the variational gap of this ELBO, which reduces to the KL divergence between the forward diffusion density and the reverse diffusion density up to a constant when $t_{\tilde{0}} \to 0$. This bound can be tightened by choosing $t_{\tilde{0}}$ closer to zero. This bound is fairly tight in practice when $t_{\tilde{0}}$ is small, and we performed an empirical analysis on a simple test case (see Appendix~ \ref{a:gap_assesment}).

\subsection{Improving Sampling Efficiency and Sample Quality}
\label{section:fast_sampling_trick}

Lastly, we introduce two techniques that can be applied to improve the efficiency of forward diffusion sampling during training, or improve sample quality post-training, which are both detailed in Appendix. The sampling strategy for the forward diffusion process presented in Section \ref{section:score_matching_training_K} requires drawing samples from $k-1$ Jacobi diffusion processes for $k$-category data, which can be demanding when $k$ is high. In Appendix~\ref{a:fast_sampling}, we describe a strategy to simplify sampling, needing to effectively sample from only one univariate Jacobi diffusion process. 

The second technique is designed to improve sample quality. Comparing to unbiasedly sampling from the learned model distribution, it is often desirable to sample near the high probability density regions, which often corresponds to higher quality samples in suitable applications. In Appendix~\ref{a:sampling_trick} we propose a simple technique, \emph{time-dilation}, applicable to reverse diffusion sampling without modifying the score model, when a flat distribution such as the flat Dirichlet distribution is the stationary distribution. In Appendix~\ref{a:comp_improve_sample}, we compare sample quality obtained by time dilation with other sampling strategies which reported to improve sample quality~\citep{song2020score}.

\section{Results}

\subsection{Implementation Notes of Dirichlet Diffusion Score Model}
Sampling from Jacobi diffusion processes is more expensive than commonly used SDEs with Gaussian stationary distributions \cite{song2020score}, as we need an SDE sampler such as Euler-Maruyama sampler. However, we only need to generate samples from two starting points, 0 and 1, for any categorical data. Hence, we can pre-sample a dictionary of diffused samples at different time points $t$ and sample from the dictionary during training time. Similarly, the log transition density function gradient at the samples can also be precomputed. This approach allows efficient training with little additional overhead. We discuss additional implementation details in the Appendix \ref{a:implementaion_notes}.

\subsection{Application to Binarized MNIST}
We first applied the method to a benchmark dataset for generative modeling, the binarized MNIST dataset, and obtained competitive performance (Table \ref{mnist_performance_table}). More details of all applications are included in Appendix \ref{a:application_details}. Examples of samples are shown in Appendix \ref{a:mnist_examples}.

\begin{table}[h!]
\vskip -0.15in
\caption{Binarized MNIST benchmark performance}
\label{mnist_performance_table}
\vskip 0.1in
\begin{center}
\begin{small}
\begin{tabular}{ll}
\toprule
Method & NLL (nats) $\downarrow$ \\
\midrule
\textbf{DDSM} (ours) & $ 78.04 \pm 0.37$ \\
 CR-VAE & $ \textbf{76.93}$ \\
Locally Masked PixelCNN & $77.58$ \\
PixelRNN & $	79.20	$ \\
PixelCNN & $		81.30		$ \\
EoNADE & $		84.68		$ \\
MADE & $	86.43	$ \\
NADE & $	88.33	$ \\

\bottomrule
\end{tabular}
\end{small}
\end{center}
\vskip -0.1in
\end{table}

\subsection{Sudoku Generation as a Constraint Satisfying Generation Test}

\begin{figure}[!t]
\begin{center}
\centerline{\includegraphics[width=\columnwidth]{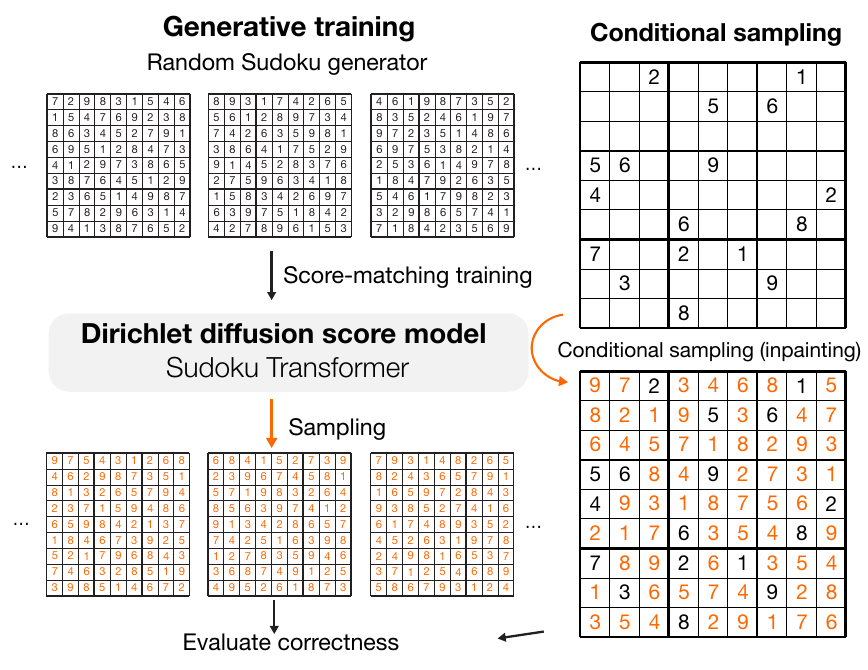}}
\vskip -0.1in
\caption{Sudoku generation and solving through generative modeling with diffusion model, as a test for constraint satisfaction. }
\label{fig:Sudoku}
\end{center}
\vskip -0.35in
\end{figure}

To test the ability to generate highly structured data that satisfy hard constraints, we applied our method to the problem of generating and solving Sudoku. This problem has not been solved through generative modeling to the best of our knowledge. 

For training, we used a Sudoku generator to continuously produce Sudoku puzzles. A fully-filled Sudoku puzzle can be encoded with 81 categorical variables with the number of categories $k=9$. The model architecture we adopt is based on a 20-block transformer architecture with a relative positional encoding designed for the Sudoku problem. 

Specifically, a Sudoku puzzle is represented by a set of 81 elements and a binary relative positional encoding that is 27 dimensional (81 x 81 x 27), corresponding to whether two elements are of the same row, same column, or the same 3x3 block. The relative positional encoding is transformed by a linear layer and added to the transformer's attention prior to the softmax.

The generative capability of the model is evaluated by the percentage of generated Sudoku that is correctly filled (Table \ref{Sudoku-performance-table}). Only whether a Sudoku is completely correct is considered, with no partial credit given. Applying the time-dilation technique (Appendix \ref{a:time_dilation}) to drive samples toward high-density areas improved sample quality to up to $100\%$. In contrast, the heuristic algorithm for generating of Sudoku that we used to generate training data, has only $0.31\%$ success rate. Even though no previous generative modeling 
 approach has been applied to Sudoku, we also trained the Sudoku Transformer with Bit Diffusion~\cite{chen2022analog} and D3PM-uniform/Multinomial Diffusion~\cite{hoogeboom2021argmax,austin2021structured}, using the same model architectures. DDSM with time dilation achieved the best performance in comparison with these methods (Appendix \ref{a:sudoku_performance_comparison}).

Interestingly, similar to prior observations on image generation quality \cite{song2020score}, we also observe a trade-off between Sudoku-solving ability and computational budget, with improved Sudoku-generation accuracy using a higher number of sampling steps and time-dilation.

\begin{table}[t]
\caption{Sudoku generation and solving accuracies for \emph{single samples} with DDSM. With \emph{multiple samples}, all Sudoku puzzles we tested were solved. See Appendix \ref{a:sudoku_performance_comparison} for comparison with baseline diffusion methods.}
\label{Sudoku-performance-table}
\begin{center}
\begin{small}
\begin{tabular}{lcc}
\toprule
Task & Time dilation & Accuracy (\%) \\
\midrule
Generation & $\textbf{8x}$ & $\mathbf{100}$\\
& 4x & $99.88 \pm 0.06$\\
 & 2x & $98.87 \pm 0.16$ \\
 & 1x & $95.08 \pm 0.46$\\
 \multicolumn{2}{l}{(Heuristic algorithm baseline)} & 0.31 \\
\hline
Solving & $\textbf{8x}$ & $\mathbf{98.26 \pm 0.18}$ \\
 & 4x  & $97.54 \pm 0.18$ \\
 & 2x  & $96.45 \pm 0.32$ \\
 & 1x  & $93.85 \pm 0.42$ \\

\bottomrule
\end{tabular}
\end{small}
\end{center}
\vskip -0.3in
\end{table}

\subsection{Solving Sudoku via Conditional Generation}

We applied the Sudoku generative SDE model to solving Sudoku puzzles by a conditional generation with the inpainting method \cite{song2020score} of clamping entries to the given clues of the puzzle. We evaluated the model on an easy Sudoku dataset with 36 clues on average \cite{Wang2019-wt} and a hard Sudoku dataset with minimally 17 clues \cite{Palm2018-recurrent}, which is the minimum number of clues possible for Sudoku \cite{McGuire2014-oy}.

Even though no additional training is done for solving Sudoku, the generative SDE model trained with DDSM solved most puzzles in the easy dataset with a single sample (Table \ref{Sudoku-performance-table}). In contrast, both models trained with Bit Diffusion and D3PM-uniform have difficulties in the Sudoku solving task, with only $<10\%$ of easy puzzles solved with a single sample.

The Sudoku-solving performance of the DDSM model can be further improved by increasing time-dilation, and a single sample solves $99.4\%$ of the easy dataset and $42.4\%$ of the hard dataset (128x time-dilation). When multiple samples are allowed, the  model can solve 100\% of all puzzles. The number of samples required to solve a Sudoku puzzle significantly increases with lower than 25 clues, with about 2.3x increase per one fewer clue given which is still significantly better than random guesses (Appendix \ref{a:hard_sudoku}). This allows us to solve 100\% hard Sudoku puzzles in the dataset. Neither models trained with Bit Diffusion nor D3PM-uniform have the ability to solve hard puzzles. Previous state-of-the-art supervised models SATNET~\cite{Wang2019-wt} and Recurrent relational network~\cite{Palm2018-recurrent} can solve most but not all of them (see Appendix \ref{a:hard_sudoku} for more discussion).


\subsection{Generation of Promoter DNA Sequences}

\begin{figure*}[!ht]
\begin{center}
\centerline{\includegraphics[width=\textwidth]{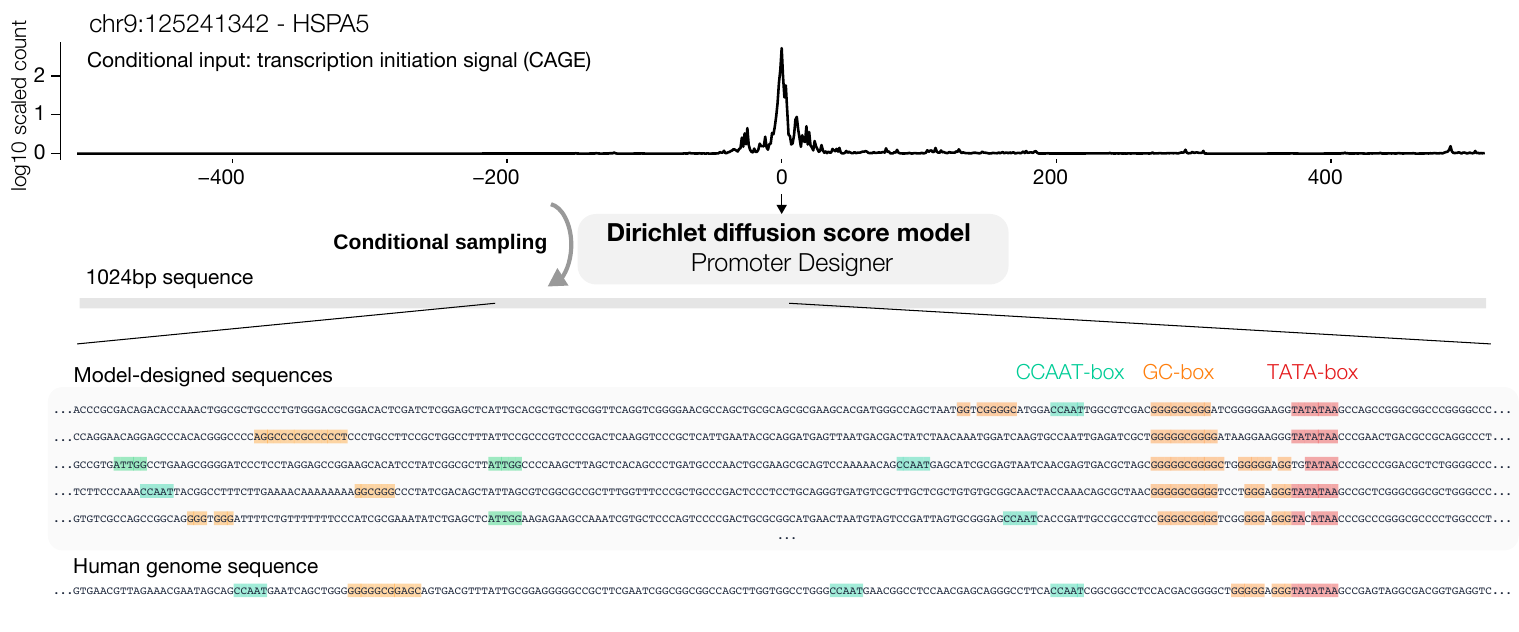}}
\vskip -0.2in
\caption{Design human gene promoter sequences with conditional Dirichlet diffusion score model trained to generate sequences from transcriptional initiation signal profile. Example model-designed sequences based on a transcriptional initiation signal profile of a test set promoter are shown. The corresponding human genome sequence is shown in comparison. Known promoter motifs are annotated in the sequences. We provide an introduction for readers unfamiliar with this topic in Appendix \ref{a:intro}. }
\label{fig:promoter1}
\end{center}
\vskip -0.35in
\end{figure*}

Finally, we applied the method to designing human promoter DNA sequences (Figure \ref{fig:promoter1}). Promoters are key DNA sequences that drive the transcription of genes and partially determine gene expression levels. Designing promoter sequences can have broad applications in biomedical research and bioengineering applications, such as controlling synthetic gene expression. Human promoter sequences are known to be highly diverse and rules that determine promoter sequence activity are not fully understood \cite{Wang2017-yr}. Thus it is an ideal problem to be addressed through deep generative modeling. No prior computational approach for designing human promoter sequences exists to our knowledge.

To enable human promoter sequence design, we trained a conditional Dirichlet diffusion score model to perform conditional generation of promoter sequences using transcription initiation signal profile as an additional input to the score model (Figure \ref{fig:promoter1}). Transcription initiation signal profiles reflect the transcription initiation activity at every sequence position and are obtained from CAGE experiments (\citealt{Forrest2014-qf}). The conditional generation model allows controlling the transcription initiation signal profile produced by the sequence, including controlling the expression level. We constructed the human promoter sequence dataset containing 100,000 promoter sequences and corresponding transcription initiation signal profiles, with each sequence 1024 basepairs long and centered at the annotated transcription start site position \cite{Hon2017-ld}. This set of promoters spans the whole range of human promoter activity levels from highly expressed protein-coding gene promoters to ncRNA gene promoters with very low expression. 

\begin{figure*}[!ht]
\begin{center}
\centerline{\includegraphics[width=\textwidth]{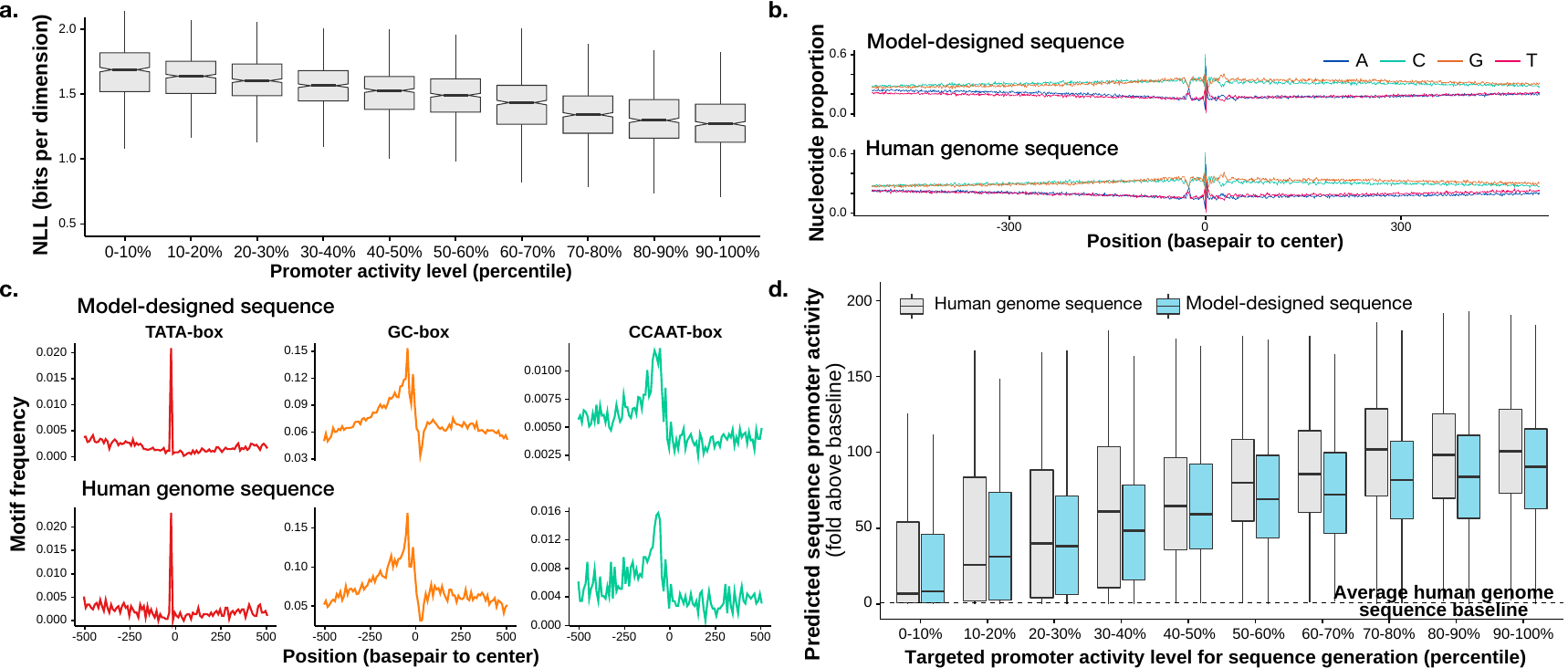}}
\caption{Performance of promoter design diffusion sequences model. a) NLL evaluated on test chromosomes 8 and 9, grouped by the expression level of the promoter (90-100\% percentile being most expressed). (b-c) Model-designed sequences have comparable position-specific nucleotide composition (b) and motif location distribution (c). d) Designed sequences (blue) are compared with human genome sequences (gray) using promoter activity predicted from sequence by Sei \cite{Chen2022-bg}. Generated sequences are grouped by the targeted promoter activity level (x-axis). Y-axis shows predicted promoter activity (average H3K4me3 prediction across cell types), divided by baseline prediction for average genomic sequences. }
\label{fig:promoter2}
\end{center}
\vskip -0.35in
\end{figure*}

\subsection{Evaluation of Designed Promoter DNA Sequences}

With a custom score-model architecture that we call Promoter Designer, the generative model obtained a conditional NLL estimate of $\le 1.32$ bits per basepair for promoters that are from test set chromosomes and among the top 10,000 promoters, whereas simple baselines using position-specific base composition achieves only $1.92$ bits. Promoter sequences with higher activity levels also obtained better NLL estimates (Figure \ref{fig:promoter2}a), in line with the expectation that sequences of high-activity promoters are less random compared to low-activity promoters. Multiple sequence samples conditioned on the same transcriptional initiation signal profile are typically diverse (Figure \ref{fig:promoter1}) while sharing similar characteristics. The generated sequences return no hits when compared with the human genome using BLAST \cite{Morgulis2008-kf}, thus the model does not simply memorize human genomic sequences.

The sequence samples are observed to contain highly similar properties as promoter sequences from the human genome (Figure \ref{fig:promoter2}b-d), such as position-specific nucleotide composition relative to the transcription start site (Figure \ref{fig:promoter2}b) as well as distribution of known promoter related motifs (Figure \ref{fig:promoter2}c). 

\begin{table}[t]
\caption{Promoter design performance comparison for different models. We trained all models with the same Promoter Designer architecture and the same early stopping criterion for this comparison.}
\label{table:promoter}
\begin{center}
\begin{small}
\begin{tabular}{ll}
\toprule
Model & SP-MSE $\downarrow$ \\ 
\midrule
\textbf{DDSM (time dilation 4x)} & \textbf{0.0334} \\ 
DDSM (time dilation 2x) & 0.0348 \\ 
DDSM (time dilation 1x) & 0.0363 \\ 
D3PM-uniform / Multinomial Diffusion & 0.0375 \\ 
Bit Diffusion (one-hot encoding) & 0.0395 \\
Bit Diffusion (bit-encoding) & 0.0414 \\ 
\bottomrule
\end{tabular}
\end{small}
\end{center}
\vskip -0.3in
\end{table}

To evaluate whether the generated sequences recapitulated more complex sequence rules of promoter activity, we applied a published deep learning sequence model Sei that can predict active promoter from sequence (based on chromatin mark H3K4me3 predictions) \cite{Chen2022-bg}, the generated sequence has comparable predicted promoter activity with the human genome sequence, for both high and low activity promoters (Figure \ref{fig:promoter2}d). Applying time-dilation further increased predicted promoter activity (Appendix \ref{a:promoter_seq_time_dilated}).

We also trained models with baseline discrete data diffusion approaches. The evaluation is based on comparing generated sequences and human genome promoter sequences (ground truth) on the test chromosomes similar to Figure 4c. The metric SP-MSE is the MSE between the predicted promoter activity of generated sequences and human genome sequences (lower is better). Our model trained with DDSM outperforms models trained with baseline approaches (Table \ref{table:promoter}).

\section{Related Work}
Diffusion models were first proposed in~\citealt{sohl2015deep,ho_denosing}, including their first application to discrete data using a binomial diffusion process for a binary dataset~\cite{sohl2015deep}. \citealt{song2020score} proposed the score-based generative SDE diffusion model framework with continuous time for continuous data. Recent works for generalizing diffusion models to discrete data have mostly considered discrete time~\cite{hoogeboom2021argmax,austin2021structured}. More recent works~\cite{campbell2022continuous, sun2022score} proposed continuous-time approaches for discrete-space diffusion based on a continuous-time Markov chain formulation. Another direction of work is to apply existing continuous-time continuous-space diffusion approach to discrete data encodings, such as bit encoding~\cite{chen2022analog} or word embedding~\cite{Li2022-vh}, and quantize the continuous samples. Concurrent to this work, \citealt{richemond2022categorical} proposed to use $k$ independent Cox-Ingersoll-Ross (CIR) processes as a forward SDE for which the stationary distribution is Gamma distribution. The authors proposed to normalize the $k$-dimensional vector obtained by the CIR processes to unit sum, and the stationary distribution is k independent gamma distributions. In addition, \citealt{lou2023reflected} introduced reflected SDEs to score-based diffusion model, which can be applied to restrict the diffusion to the probability simplex.

Latent diffusion models also allow generative modeling of discrete data by modeling only the distribution of continuous latent variable with diffusion \cite{vahdat2021score,Lovelace2022-nv}. While the reverse diffusion process in our approach can also be interpreted as a latent variable that emits discrete data (Section \ref{section: discretelikelihood}), the relationship between latent and discrete variables is fixed rather than learned.

On generative modeling for biological sequence design, deep generative models have been recently applied to DNA sequence design \cite{Killoran2017-hl,wang2020synthetic, Zrimec2022-sk}, even though no diffusion model has been developed. No prior method exists for designing human promoter sequences to our knowledge. On the protein sequence design problem, several deep generative models have been developed to generate sequences conditioned on protein structure \cite{Ingraham2019-nc,Dauparas2022-dk}, and diffusion models have been applied to generate protein structure \cite{Trippe2022-km,Watson2022-lq,Lee2022-kt}. Recently works also jointly generate structure and sequence with diffusion \cite{Luo2022-ro,Anand2022-vv}.

Our contribution is to propose the approach for discrete data modeling with continuous-time SDE diffusion in probability simplex space, and applied this approach to develop the first method for human promoter sequence design and a novel application to generative modeling of Sudoku. All existing works using score-based generative SDEs are based on diffusion processes that converge to Gaussian stationary distributions, and here we expand the generative SDE toolkit to include ones that converge to Dirichlet stationary distribution. In addition, we propose a simple and easily applicable technique, time-dilation, to improve sample quality.

\section{Discussion}

We provided a continuous-time Dirichlet diffusion score model framework (DDSM), for generative modeling of biological sequences, which can also be used for other types of discrete data. The approach also provides a plug-in substitute for Gaussian stationary distribution SDEs for discrete variables and expands the toolkit of generative diffusion model.

The size of the continuous diffusion state scales linearly with the number of categories and thus may not directly scale to a very high number of categories due to memory and computational constraints. A potential way to address this issue is to use bit encoding or hierarchical encoding schemes that can reduce the encoding dimensions required down to $\log_2(C)$, where $C$ is the number of categories. However, the current approach is ideal for applications in modeling biological sequences, such as DNA and RNA sequences (4 bases) and protein sequences (20 amino acid residues), as well as other data that can be encoded with a moderate number of categories. 

We are encouraged by the promising results in designing human promoter sequences and the strong constraint satisfaction capability demonstrated in generating and solving sudoku, and look forward to further development in biological sequence design based on this approach.

\section*{Acknowledgements}
This work is supported by Cancer Prevention and Research Institute of Texas grant RR190071, NIH grant DP2GM146336, and the UT Southwestern Endowed Scholars Program.

\bibliographystyle{icml2021}
\bibliography{main.bib}


\newpage
\appendix
\onecolumn
\section{Supplemental Information for Dirichlet Diffusion Score Model} 
\subsection{Transition Density Function of Jacobi Diffusion Process}

For Jacobi diffusion process
\begin{equation*}
\mathrm{d}\mathbf{x} = \frac{s}{2}[a(1-\mathbf{x})-b\mathbf{x}]\mathrm{d}t + \sqrt{s\mathbf{x}(1-\mathbf{x})} \mathrm{d}\mathbf{w},
\end{equation*}
the transition density function is 
\begin{equation*} 
\begin{aligned}
p_{a,b}(x^t|x^0) = \mathcal{B}_{a,b}\left(x^t\right) \sum_{n=0}^{\infty}\frac{e^{\lambda_n t}}{d_n} R_n^{(a,b)}\left(x^0\right)R_n^{(a,b)}\left(x^t\right) = \mathcal{B}_{a,b}\left(x^t\right) \left(1+\sum_{n=1}^{\infty}\frac{e^{\lambda_n t}}{d_n} R_n^{(a,b)}\left(x^0\right)R_n^{(a,b)}\left(x^t\right)\right),
\end{aligned}
\end{equation*}
where $\mathcal{B}_{a,b}(x_t)$ is the $\textbf{Beta}(a,b)$ density, 
$$R_n^{(a, b)}(x)= P_n^{(b-1,a-1)}(2x-1)$$ denotes the $n$-th order modified Jacobi polynomial of order $n$.  $$d_n=\frac{a_{(n)} b_{(n)} }{(a+b)_{(n-1)}(2n+a+b-1)n!}$$ is the $n$-th order constant. 

$$a_{(n)}=\frac{\Gamma(a+n)}{\Gamma(a)}=\prod_{k=0}^{n-1}(a+k)$$ 
denotes rising factorial also known as the Pochhammer symbol and $$P_n^{(\alpha,\beta)}(x) = \frac{\Gamma(\alpha+n+1}{n!\Gamma(\alpha+\beta+n+1)}\sum\limits_{m=0}^n {n \choose m}\frac{\Gamma(\alpha+\beta+n+m+1)}{\Gamma(\alpha+m+1)}\left(\frac{x-1}{2}\right)^m$$ 
is the Jacobi polynomial. 
$R_n^{(a, b)}(x)$ are eigenfunctions of the generator of the Jacobi diffusion process \cite{steinrucken2013explicit,griffiths2010diffusion}. The corresponding eigenvalues are $\lambda_n=-\frac{1}{2}sn(n-1+a+b)$. 

\subsection{Choosing Speed Factor $s$ in Jacobi Diffusion Processes for Dirichlet Diffusion Score Model}\label{a:speed_factor}
Let us recall Jacobi diffusion process from Section~\ref{section:jacobi_diff_proc}: 
$$\mathrm{d}\mathbf{x} = \frac{s}{2}[a(1-\mathbf{x})-b\mathbf{x}]\mathrm{d}t + \sqrt{s\mathbf{x}(1-\mathbf{x})} \mathrm{d}\mathbf{w}.$$
The choice of $s$ affects the convergence speed of Jacobi diffusion process while having no effect on the stationary distribution. In other words, the diffusion process converges to stationary distribution faster with higher $s$. 

For modeling $2$-category data with univariate Jacobi diffusion processes, the choice of $s$ only affects the appropriate selection of maximum time in the diffusion model, which should be chosen inversely proportional to $s$. For modeling $k$-category data with multivariate Jacobi diffusion process by stick-breaking construction, the selection of $s$ can affect the relative convergence speed for each independent univariate Jacobi forward diffusion processes.

We propose two options of $s$:

\begin{enumerate}[(i)]
    \item The first option, $s=1$, was empirically observed to achieve uniform convergence speed in $\mathbf{x}$ space across dimensions corresponding to different categories. This also generates samples that are nearly identical to fast sampling strategies described in Section \ref{a:fast_sampling}. Choosing $s=1$ generally requires selecting a lower maximum time for more categories, to compensate for the faster convergence with higher $k$.
    \item The second option, $s=\frac{2}{a+b}$, ensures uniform converge speed in the $\mathbf{v}$ space. This is motivated by choosing the first eigenvalue of the transition density function $\lambda_1=-\frac{1}{2}s(a+b)$ to be equal across Jacobi diffusion processes. It also allows conveniently keeping a fixed maximum time for diffusion modeling (e.g. $4$), regardless of the number of categories $k$ in the data or the $a,b$ parameters of the Jacobi diffusion.
\end{enumerate}

We tested both variants of $s$ and obtained good empirical results. Hence, these choices are often interchangeable. We used $s=1$ for Sudoku generation and $s=\frac{2}{a+b}$ for promoter sequence design. The binarized MNIST application uses only the univariate Jacobi diffusion process where the two options are equal. We recommend comparing the choices on specific applications.

\subsection{Weighting Function for Score Matching Loss Invariant to Change-of-Variable}\label{a:weighting_function}

Here we will show that the proposed weighted score matching loss is invariant to change-of-variable for any SDE. To first show that the unweighted score matching loss is not invariant to change-of-variable, we consider change of variable by any bijective, differentiable function $\mathbf{x}=h(\mathbf{v})$. Applying the change-of-variable equation for probability density function, the unweighted score matching loss
\begin{equation*}
\begin{split}
\bigg\| 
\frac{\partial{\log p_{\mathbf{x}}(\mathbf{x})}}{\partial \mathbf{x}}
- \frac{\partial{\log q_{\mathbf{x}}(\mathbf{x})}}{\partial \mathbf{x}}
\bigg\|_2^2  
= \bigg\| \left(\frac{\partial{\log p_{\mathbf{v}}(\mathbf{v})}}{\partial \mathbf{v}} - \frac{\partial{\log q_{\mathbf{v}}(\mathbf{v})}}{\partial \mathbf{v}}\right) \frac{\partial \mathbf{v}}{\partial \mathbf{x}} \bigg\|_2^2 
 \end{split}
\end{equation*} 
is not invariant to the change of variable due to the extra $\frac{\partial \mathbf{v}}{\partial \mathbf{x}}$ term. We now show that the loss function Equation \ref{eq:weightedloss} (also shown below) is invariant to change of variable $\mathbf{x}= h(\mathbf{v}, t)$ (where $\mathbf{x} = h(\mathbf{v})$ is a special case).

\begin{equation*} 
\begin{split} 
L(\mathbf{v}, t) = \Biggl\| \frac{\partial{\log p_{\mathbf{v}}(\mathbf{v})}}{\partial \mathbf{v}} &- \frac{\partial{\log q_{\mathbf{v}}(\mathbf{v})}}{\partial \mathbf{v}} \Biggr\|_{\mathbf{G}\mathbf{G}^T}^2 \\
&= 
\left( \frac{\partial{\log p_{\mathbf{v}}(\mathbf{v})}}{\partial \mathbf{v}} - \frac{\partial{\log q_{\mathbf{v}}(\mathbf{v})}}{\partial \mathbf{v}} \right) \mathbf{G}(\mathbf{v}, t) 
\mathbf{G}(\mathbf{v}, t)^T \left( \frac{\partial{\log p_{\mathbf{v}}(\mathbf{v})}}{\partial \mathbf{v}} - \frac{\partial{\log q_{\mathbf{v}}(\mathbf{v})}}{\partial \mathbf{v}} \right)^T.
\end{split}
\end{equation*}

For Ito diffusion process
$$\mathrm{d} \mathbf{v} = \mathbf{f}(\mathbf{v}, t)\mathrm{d}t +\mathbf{G}(\mathbf{v}, t)\mathrm{d}\mathbf{w},$$
change-of-variable to $\mathbf{v}$ gives the following Ito diffusion process due to Ito's lemma
\begin{equation*}
\begin{aligned}
\mathrm{d} \mathbf{x} =\Biggl\{\frac{\partial \mathbf{x}}{\partial t}+
\frac{\partial \mathbf{x}}{\partial \mathbf{v}} \mathbf{f}(\mathbf{v},t)+\frac{1}{2} \operatorname{Tr}\left[\mathbf{G}(\mathbf{v}, t)^{T}\left(\frac{\partial^2 \mathbf{x}}{\partial \mathbf{v}^2} \right) \mathbf{G}(\mathbf{v}, t)\right]\Biggr\} \mathrm{d} t
&+\frac{\partial \mathbf{x}}{\partial \mathbf{v}} \mathbf{G}(\mathbf{v}, t) \mathrm{d} \mathbf{w}.
\end{aligned}
\end{equation*}
Thus, $\mathbf{G}(\mathbf{x}, t) = \frac{\partial \mathbf{x}}{\partial \mathbf{v}} \mathbf{G}(\mathbf{v}, t)$. Plugging this in equation \ref{eq:weightedloss}, we see that $L(\mathbf{x},t) = L(\mathbf{v},t)$. 

\subsection{Improving Sampling Efficiency for Data with High Number of Categories}\label{a:fast_sampling}

The sampling strategy for the forward diffusion process presented in Section \ref{section:score_matching_training_K} requires drawing samples from $k-1$ Jacobi diffusion processes for $k$-category data, which can be demanding when $k$ is high. Here we describe a strategy to accelerate sampling, needing to effectively sample from only one univariate Jacobi diffusion process.

We assume that the stationary distribution is the flat Dirichlet distribution \textbf{Dir}$(1,1,\dots,1)$. We reorder the sequence of stick-breaking construction, starting from the dimension with value 1 in one-hot encoding first, which allows us to initialize with $v_1=1$ and diffuse with Jacobi diffusion $(a=1, b=k-1)$. This reordering allows all other $v_i$ values to be drawn without sampling from Jacobi diffusion as they can be drawn directly from their stationary distribution $v_i \sim \textbf{Beta}(1,k-i)$. The samples will be converted to $x$ space and reordered back to the original order. Reordering does not change the initial or stationary distribution of the diffusion process. 

This sampling strategy leads to exactly $k-1$ fold speed up and lower memory consumption during precomputation of the samples and scores. During training time, we also observed it to be faster than the regular sampling (for example, $2.08$ vs $2.87$ ms for the fast sampling method vs regular sampling method for $10,000$ dimensions).

Empirically, the fast sampling method generates nearly identical samples as the original multivariate Jacobi diffusion process by stick-breaking construction with all $s$ factors set to constant. We find it hard to detect any noticeable decrease in sample quality or in log likelihood, showing that the effect is likely very small (on a synthetic data set, we observed $2.08$ bits /dim using the model trained with the fast sampling strategy, whereas the ground truth optimal likelihood is $2$ bits/dim). For example, our Sudoku model is trained with the fast sampling strategy and achieves perfect accuracy in Sudoku generation (see Appendix~\ref{a:sudoku_design_details} for more details).


\subsection{Improving Sample Quality by Biasing Reverse Diffusion Toward High-Density Areas}\label{a:sampling_trick}
Compared to unbiasedly sampling from the learned model distribution, it is often desirable to sample near the high probability density regions. These regions often correspond to higher-quality samples. We propose a simple technique applied to reverse diffusion sampling without modifying the score model when a flat distribution is the stationary distribution (e.g., the flat Dirichlet distribution).

There are two equivalent modifications of the reverse diffusion process during sampling: 
\begin{enumerate}[(i)]
\item increasing the maximum time of reverse diffusion by a factor of $k$ while querying the score model (and diffusion coefficient if it is time-dependent) with time proportionally scaled back to the original; or 
\item accelerating the reverse SDE by a factor $c$ while keeping the maximum time and score model of reverse diffusion unchanged. More specifically, we have
\begin{equation*}
\begin{aligned}
\mathrm{d} \mathbf{x}&=c\biggl\{\mathbf{f}(\mathbf{x}, t)-\nabla \cdot\left[\mathbf{G}(\mathbf{x}, t) \mathbf{G}(\mathbf{x}, t)^{\top}\right]-\mathbf{G}(\mathbf{x}, t) \mathbf{G}(\mathbf{x}, t)^{\top} \nabla_{\mathbf{x}} \log p_{t}(\mathbf{x})\biggr\} \mathrm{d} t+\sqrt{c}\mathbf{G}(\mathbf{x}, t) \mathrm{d} \overline{\mathbf{w}}.
\end{aligned}
\end{equation*}
 \end{enumerate}

We call these techniques \emph{time dilation}. 

The modified SDE can be considered a reverse-time SDE for the forward diffusion accelerated by $c$. However, the score function is biased upward since it is estimated from the original forward diffusion which converges to stationary distribution slower. Thus, the samples tend to lie in higher-density areas. Since time dilation drives samples toward high-density areas, sometimes it is only desired to do locally instead of globally. Therefore, we find it effective to achieve so by starting time-dilation only at later stages of reverse diffusion.

It is also important to note that time dilation does not increase the time complexity of the sampling unless the number of sampling steps is increased. For example, $128$x  time dilation on $3200$ time steps involves only $3200$ evaluations. We generally recommend increasing the number of sampling steps proportional to the time dilation.

Overall, time dilation represents a general technique that can be applied to any continuous-time diffusion model.

\section{Implementation Notes of Dirichlet Diffusion Score Model (continued)}\label{a:implementaion_notes}
\subsection{Implementation of the Jacobi Diffusion Transition Density Function}
\label{a:implementaion_notes_diff}
We compute Jacobi polynomials using dynamic programming. Hence, the time complexity of the algorithm is linear with respect to the number of terms in the Jacobi diffusion process. Table~\ref{table:jacobi} provides information about the computation time of the Jacobi diffusion density function with up to 1000 terms in eigendecomposition for the input of $10000$ dimensions.

\begin{table}[h]
\centering
\begin{tabular}{|l|c|c|c|c|c|}
\hline
Order of Jacobi Polynomials & 10 & 50 & 100 & 500 & 1000 \\ \hline
Runtime - Pytorch (ms) & 0.27 & 1.33 & 2.33 & 9.33 & 19.8 \\ \hline
\end{tabular}
\caption{Runtime of Jacobi diffusion density function computation on PyTorch 1.10.1.}
\label{table:jacobi}
\end{table}

By choosing the number of terms for computing the Jacobi diffusion density function, one seeks a trade-off between running time and numerical issues. Table~\ref{table:gap} contains the relative error for the log gradient of the Jacobi diffusion density function ($a=1$, $b=3$) for different time points, averaged over $1000$ samples for each time point, using scores evaluated with $10000$ terms as the ground truth. Table~\ref{table:gap} shows that $1000$ terms are sufficient for precisely computing the score for $t=0.001$, $100$ terms are sufficient for $t=0.01$, and $20$ terms are sufficient for $t=0.1$.

\begin{table}[h]
\centering
\begin{tabular}{|c|c|c|c|c|}
\hline
\# of terms & t=0.001 & t=0.01 & t=0.1 & t=1 \\ \hline
10 & 11.2 & 14.0 & 1.1 & 0 \\ \hline
20 & 10.8 & 7.4 & 0 & 0 \\ \hline
50 & 7.5 & 0 & 0 & 0 \\ \hline
100 & 0.2 & 0 & 0 & 0 \\ \hline
200 & 0 & 0 & 0 & 0 \\ \hline
500 & 0 & 0 & 0 & 0 \\ \hline
1000 & 0 & 0 & 0 & 0 \\ \hline
\end{tabular}
\caption{The relative error for the log gradient of Jacobi diffusion density function for different time points, averaged over $1000$ samples for each time point.}
\label{table:gap}
\end{table}

We decided to be conservative and chose to use 1000 terms for all our experiments. For numerical accuracy, we recommend evaluating Jacobi diffusion with eigendecomposition up to the 1000th term as well, with double precision floating point arithmetic, for time greater or equal to $0.001$. It is also recommended to perform the stick-breaking transform and its inverse transform in double precision. Using double precision in these steps generally incurs negligible performance costs and noticeably improves numerical accuracy.

As described in the main text, we can presample a dictionary of diffused samples for each independent Jacobi diffusion process at uniformly spaced time points to allow efficient training. In the next section, we will discuss it in more detail.



\subsection{Efficient Sampling and Score Computation from Jacobi Diffusion Processes}
The computational complexity of both sampling and score computation from diffusion process for $k$ category data is $O(n)$, where $n$ is the dimensions of the input (e.g., a length of $1000$ sequence with $4$ categories has $4000$ dimensions). We note that our diffusion process is specifically designed to be linear with respect to the number of categories, whereas other choices of multivariate diffusion processes with Dirichlet stationary distribution may have quadratic complexity (e.g. multivariate Wright-Fisher diffusion) with respect to the number of categories.

While our diffusion evaluation is more expensive than the commonly used diffusion process with Gaussian stationary distribution, all Jacobi diffusion-related computations can be precomputed prior to training and do not add to training time. This is feasible because we only need to generate samples from two starting points, 0 and 1, for any categorical data.

Thus, we can presample a dictionary of diffused samples for each independent Jacobi diffusion process at uniformly spaced time points to allow efficient training. For most applications, it suffices to sample a dictionary containing $100,000$ diffusion samples for each of $400$ uniformly spaced time points for each Jacobi diffusion process. The presampled dictionary can be saved and reused for applications using the same forward diffusion processes. This approach applies to not just the standard sampling approach but also the fast sampling approach in Section \ref{a:fast_sampling}.
 
Sampling during training is done by choosing randomly from the pre-sampled samples and scores. The sampling time is negligible compared to neural network training. Thus, training/sampling/likelihood evaluation processes have the same complexity as the previous score-based SDE diffusion model. For example, only $2.87$ms is needed for both generating a sample and its score for $10,000$ dimensions combined.

We believe that the whole process can be further optimized since only time points very close to zero would require a high number of terms to ensure numerical accuracy (see Appendix~\ref{a:implementaion_notes_diff}). These optimizations together with JIT-based speed up may eliminate the need for precomputation without slowing down the training, sampling, and likelihood computation processes.

\subsection{Importance Sampling of Time During Denoising Score-Matching Training}\label{a:time_dilation}
Importance sampling is often needed to stabilize the training of diffusion model with likelihood weighting \cite{Song2021-zl}, since the scores for the forward diffusion transition density functions are usually large when time is small. Importance sampling is thus used as a variance reduction technique that samples time non-uniformly during training. By sampling time points where the scale of the score is higher more often and down-weight the sample loss accordingly, the variance of the gradient can be reduced. We determine the importance sampling weight based on the scale of the scores at each time point observed empirically. While different choices of importance sampling weights can be used, we found $$w(t) \propto \mathbb{E}_{p_0(\mathbf{v}^0) p\left(\mathbf{v}^t \mid \mathbf{v^0}\right)}\left\|s\mathbf{v}(1-\mathbf{v}) \nabla_{\mathbf{x}} \log p\left(\mathbf{x}^t \mid \mathbf{x}^0\right)\frac{\partial \mathbf{x}}{\partial \mathbf{v}} \right\|_F,$$ where the norm is the Frobenius norm, to be a good choice for most applications.

\subsection{Score Model Design}\label{a:score_model_design}
While the architecture of the score model should be designed for the specific data type and problem. There are several aspects of score model design that are shared in common.

First, the score model should take continuous time $t$ as input, we found Gaussian Fourier projection to work well as the time embedding function following \citealt{song2020score} which in turn followed \citealt{Tancik2020-we}. 

Second, as discussed in Section \ref{a:implementaion_notes}, the scale of the score function is dependent on time, especially when $t$ is small. We can leave it to the score model to learn this time dependency or introduce a time-dependent scaling explicitly in the score model thus the model needs to learn the residual dependencies on time and input. For example, the last layer output can be multiplied with the time-dependent weight $w(t)$ in the above section with linear interpolation between time points.

\subsection{Sampling}

We used Euler Maruyama sampler and the modified version with time dilation (Section \ref{a:fast_sampling}) for all our applications for simplicity. Many improved sampling approaches have been proposed for more efficient sampling. We leave the exploration of applying these sampling approaches with Dirichlet diffusion score models for future research.

We discretize the samples by using argmax to choose the sampled category among $k$ categories, even though mapping samples to discrete samples is trivial for trained models since the samples are generally close to 1 in one category and close to 0 in all other categories. 

\subsection{Selection of the Minimum Time for Diffusion Processes}
The score of the transition density function at $t=0$ does not exist and the score at very small $t$ tends to become very large and cause numerical issues. Thus in practice, a choice of the minimum time used for training, sampling, and likelihood or ELBO evaluation is needed. With typical choices for diffusion parameters suggested in the manuscript, a minimum time of 0.01 or 0.001 is sufficient.

\subsection{Randomization of Stick-Breaking Construction Order}
We do not in general observe the order of stick-breaking construction in multivariate Jacobi diffusion to affect the model performance or samples. However, it is possible to enforce order invariance by randomizing the stick-breaking construction order during training or sampling. As the score model is formulated in $\mathbf{x}$ space, it can be converted to $\mathbf{v}$ space with any stick-breaking transform order.

\subsection{The empirical assessment on the tightness of ELBO}
\label{a:gap_assesment}
For measuring the gap, we created a simple test case with $4$ categories for which the ground truth data density is known. Table~\ref{table:ELBO} contains the measured gap between the ELBO and the ground truth likelihood. It represents an upper bound of the ELBO variational gap.

\begin{table}[h]
    \centering
    \begin{tabular}{c|c}
        \hline
        $t_{\tilde{0}}$ & Gap (bits/dim) \\ 
        \hline
        0.001 & 0.0023 \\ 
        0.002 & 0.0045 \\
        0.005 & 0.0111 \\
        0.010 & 0.0219 \\
        \hline
    \end{tabular}
    \caption{Empirical assessment of the gap between our ELBO and the ground truth likelihood. $t_{\tilde{0}}$ is a time close to $0$. }
    \label{table:ELBO}
\end{table}

We conclude that this gap should be tight enough for most applications and can be further tightened by lowering $t_{\tilde{0}}$, where $t_{\tilde{0}}$ is a time close to $0$.

\subsection{Comparison of Time Dilation Approach with Other Improving Sample Techniques}
\label{a:comp_improve_sample}
We compared the proposed time dilation approach with predictor-correct sampling using the sudoku generation task (see Appendix~\ref{a:sudoku_design_details} for more details about the sudoku experiments). Table~\ref{a:sudoku_design_details} contains results from the Predictor-Corrector sampler with the number of corrector steps being 1, 3, 7 and from the time dilation approach. Both techniques use the same number of evaluations. From Table~\ref{a:sudoku_design_details}, we conclude that time dilation approach consistently shows better performance. However, it is important to note that time dilation is a \emph{biased} sampling technique (i.e., sample preferentially from high-density areas), whereas predictor-corrector sampling is intended for unbiased sampling.
\begin{table}[h]
    \centering
    \begin{tabular}{c|c}
        \hline
        Model & Accuracy \\ 
        \hline
        \textbf{Time dilation 8x} & \textbf{100} \\ 
        Time dilation 4x & 99.88 $\pm$ 0.06 \\
        Time dilation 2x & 98.87 $\pm$ 0.16 \\
        Predictor-Corrector (7 corrector steps) & 98.86 $\pm$ 0.26 \\
        Predictor-Corrector (3 corrector steps) & 97.70 $\pm$ 0.26 \\
        Predictor-Corrector (1 corrector step) & 95.16 $\pm$ 0.47 \\
        Baseline & 95.08 $\pm$ 0.46 \\
        \hline
    \end{tabular}
    \caption{Accuracy comparison of different sampling methods.}
    \label{a:sudoku_design_details}
\end{table}


\section{Application Details}\label{a:application_details}

In this section, we provide more details about the experiments and applications.

\subsection{Binarized MNIST}
The model architecture is adopted from \citealt{ho_denosing}, and replaces the time embedding with Gaussian Fourier projection-based continuous-time embedding. The model is trained with univariate Jacobi diffusion with $s=1$, time-dependent weight-based scaling in the score model (Section \ref{a:score_model_design}), minimum time of 0.001, and maximum time of 4.

\subsection{Sudoku Generation and Solving}\label{a:sudoku_design_details}

The Sudoku training samples are fully-filled Sudokus sampled from the Sudoku generation code (\url{https://github.com/Kyubyong/sudoku/blob/master/generate_sudoku.py}, which is itself an adaptation of Sudoku generation code from \url{https://www.ocf.berkeley.edu/~arel/sudoku/main.html}). Sudoku puzzles are randomly generated puzzles by the "pluck" method of this code. We also measured the success rate of this Sudoku generation algorithm, which iteratively fills the Sudoku puzzle with numbers with no conflict, until it is no longer possible. As a baseline, the heuristic algorithm only has $0.31\%$ accuracy. 

The Sudoku transformer is a 20-block transformer architecture with the attention-bias style relative positional embedding, as described in the main text. The Sudoku transformer model is trained with the fast sampling strategy described in Section \ref{a:fast_sampling} with maximum time $1$. 

For generation and solving Sudoku puzzles, we used Euler Maruyama sampler with and without time-dilation technique for reverse diffusion sampling. $100k$ steps where $k$ is the time-dilation factor are used. To estimate the accuracy of easy and hard Sudoku puzzles with a single sample and 128x time-dilation, we used 3200 steps. To solve Sudoku with multiple samples, 8x time dilation with 200 steps was used, and we keep generating new samples until the generated sample solved the Sudoku puzzle.


\subsection{Sudoku performance comparison with baseline methods}\label{a:sudoku_performance_comparison}
\begin{table}[hbt!]
\caption{Sudoku generation and solving accuracies for \emph{single samples}, in comparison with baseline diffusion methods. We trained all models with the Sudoku transformer architecture. }
\label{Sudoku-generation-table-full}
\begin{center}
\begin{small}
\begin{tabular}{lll}
\toprule
Task & Method & Accuracy (\%) \\
\midrule
Generation & \textbf{DDSM (time dilation 8x)} & $\mathbf{100}$\\
 & DDSM (time dilation 4x) & $99.88 \pm 0.06$\\
 & DDSM (time dilation 2x) & $98.87 \pm 0.16$ \\
 & DDSM (time dilation 1x) & $95.08 \pm 0.46$\\
 & Bit Diffusion & $99.60 \pm 0.11$ \\
 & D3PM-uniform / Multinomial Diffusion & $98.90 \pm 0.18$ \\
\hline
Solving & \textbf{DDSM (time dilation 8x)} & $\mathbf{98.26 \pm 0.18}$ \\
 & DDSM (time dilation 4x)  & $97.54 \pm 0.18$ \\
 & DDSM (time dilation 2x)  & $96.45 \pm 0.32$ \\
 & DDSM (time dilation 1x)  & $93.85 \pm 0.42$ \\
 & Bit Diffusion & $7.48 \pm 0.55$ \\
 & D3PM-uniform / Multinomial Diffusion & $7.37 \pm 0.58$ \\
\bottomrule
\end{tabular}
\end{small}
\end{center}
\vskip -0.3in
\end{table}

\subsection{Human Promoter Sequence Design}\label{a:promoter_design_details}

You can refer to Appendix \ref{a:intro} if you are looking for more background information on the promoter design problem.

For the preparation of the dataset, we first obtained human TSS position annotation from the FANTOM-CAT catalog using the Level 3 (Robust) annotations. We obtained transcription initiation signal profiles measured by CAGE from the FANTOM project \cite{Forrest2014-qf}. FANTOM CAGE datasets were downloaded from \url{https://fantom.gsc.riken.jp/5/datafiles/latest/}. We averaged all CAGE profiles after applying $\log(x+1)$ transformation to obtain a robust genome-wide transcription initiation signal profile.

The human genome sequences are retrieved from hg38, with each sequence 1024 bp centered at the annotated TSS position. The sequences are retrieved on the same strand as the annotated direction of transcription of the promoter. In total, 100,000 promoters with the highest expression are retrieved. The promoters are further split into the training, validation, and test sets based on chromosomes (chr8 and 9 for the test set, chr10 for the validation set, and all other chromosomes for the training set). In the training set, we also introduce the same amount of random shift of up to $+/- 100\text{bp}$ to the sequence and transcription initiation profile simultaneously. The sequences and transcription profile profiles at centered at a location within $+/- 100\text{bp}$ distance to the annotated TSS in this case. Random shifts are only used during training as a data augmentation and regularization technique.

The Promoter Designer model has a custom-designed 1D convolutional architecture. This conditional generation model concatenates the 4-dimensional $\mathbf{x}$ input and the 1-dimensional transcription initiation signal (CAGE) profile. The training uses $s=\frac{2}{a+b}$ Jacobi diffusion processes with maximum time $4$. For sampling from the trained model, we used Euler Maruyama sampler with 100 steps.

For analysis of generated sequences and comparison with human genome sequences, we generated sequences conditioned on the test set transcription initiation profiles among the top 40,000 promoters. 5 sequences are generated for each transcription initiation profile. The human genome sequences for these test set promoters are used for comparison. For motif position distribution analysis, we used the following motifs from JASPAR database \cite{Castro-Mondragon2022-sw}: TATA-box, AC0057:TBP/ZNF:TBP; GC-box, AC0524:KLF/SP:C2H2\_ZF; CCAAT-box, AC0240:NFYA/NFYB:CBF/NF-Y. Sei model \cite{Chen2022-bg} is trained with the same validation and test holdout chromosomes as Promoter Designer, and is thus ideal for evaluation of Promoter Designer. For Sei model prediction, we padded the input sequence to 4096bp with 0.25 and averaged the prediction for all H3K4me3 targets to obtain the predicted sequence promoter activity score.

For comparison with baseline diffusion model methods, we trained all models with the same Promoter Designer architecture, including retraining the DDSM model, using the same early stopping criterion (SP-MSE on the validation set).

\newpage
\section{Binarized MNIST Generation Examples Including Time-Dilation Experiments}\label{a:mnist_examples}
\counterwithin{figure}{section}

We show below samples from the Dirichlet diffusion score model, with and without applying time-dilation (Appendix \ref{a:time_dilation}) in sampling. 

Consistent with our expectation that time-dilation produces samples biased toward high-density regions, we notice that samples with time-dilation generated more stereotypical digits. Further increasing time dilation also causes the samples to contain more \say{1}s, which is likely due to that \say{1} is slightly more common than other digits in the binarized MNIST dataset. The second most common digit \say{7} is also over-represented in highly time-dilation samples, consistent with the hypothesis that the non-uniform distribution among digits in the dataset is exaggerated by time dilation.

We get the best of both worlds i.e. increased digit quality while not biased toward overrepresented digits, but starting the time-dilation only at a later stage of reverse diffusion.

\begin{figure}[h!]
\vskip 0.2in
\begin{center}
\centerline{\includegraphics[width=0.5\textwidth]{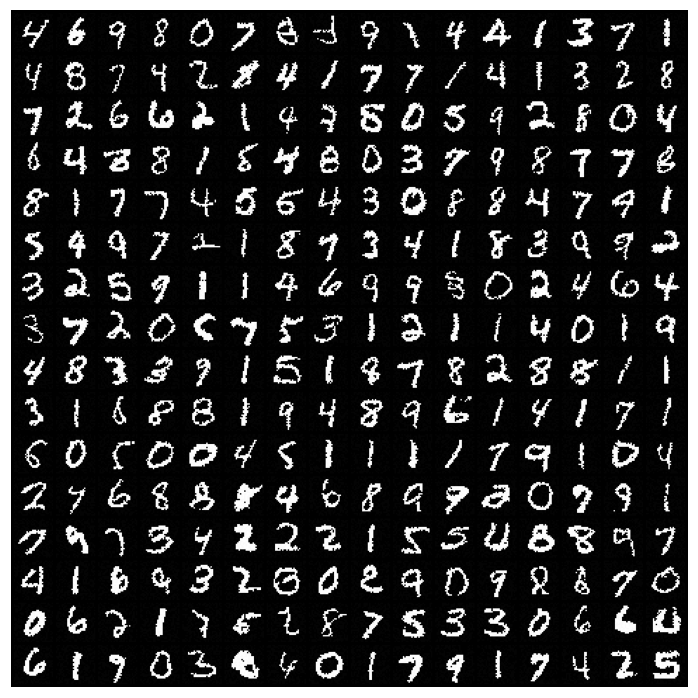}}
\caption{Binarized MNIST samples from Dirichlet diffusion score model. Euler Maruyama Sampler with 100 steps. }
\label{fig:mnist_examples}
\end{center}
\vskip -0.2in
\end{figure}

\pagebreak

\begin{figure}[h!]
\vskip 0.2in
\begin{center}
\centerline{\includegraphics[width=0.5\textwidth]{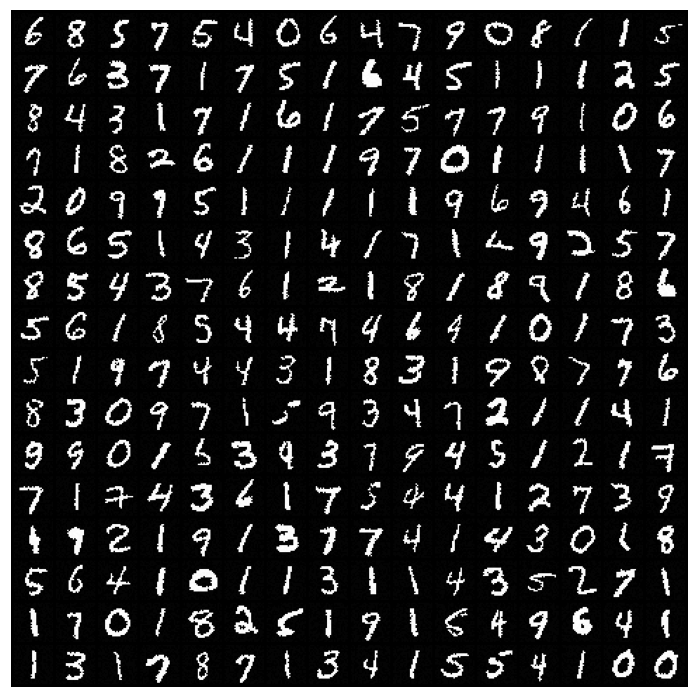}}
\caption{Binarized MNIST samples from Dirichlet diffusion score model (2x time-dilation). Euler Maruyama Sampler with 2x time-dilation and 200 steps. }
\label{fig:mnist_examples2x}
\end{center}
\vskip -0.2in
\end{figure}

\begin{figure}[h!]
\vskip 0.2in
\begin{center}
\centerline{\includegraphics[width=0.5\textwidth]{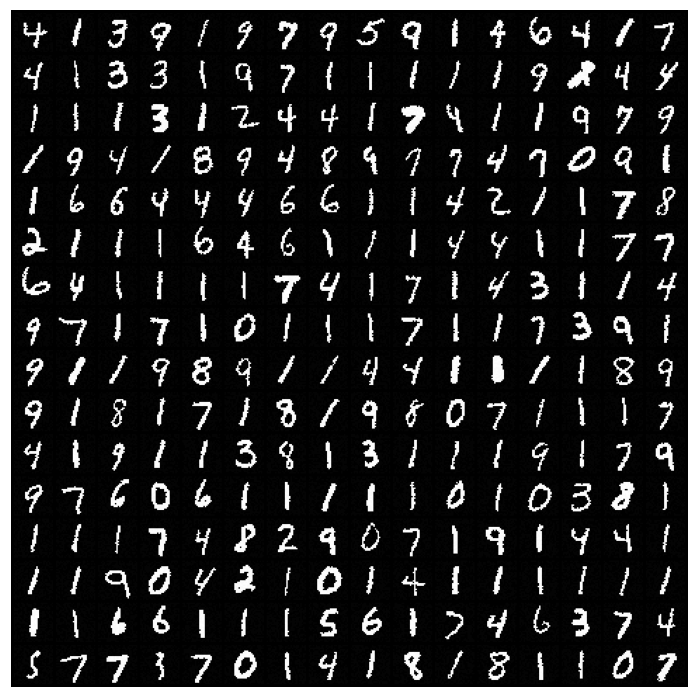}}
\caption{Binarized MNIST samples from Dirichlet diffusion score model (4x time-dilation). Euler Maruyama Sampler with 4x time-dilation and 400 steps. }

\label{fig:mnist_examples4x}
\end{center}
\vskip -0.2in
\end{figure}

\pagebreak

\begin{figure}[h!]
\vskip 0.2in
\begin{center}
\centerline{\includegraphics[width=0.5\textwidth]{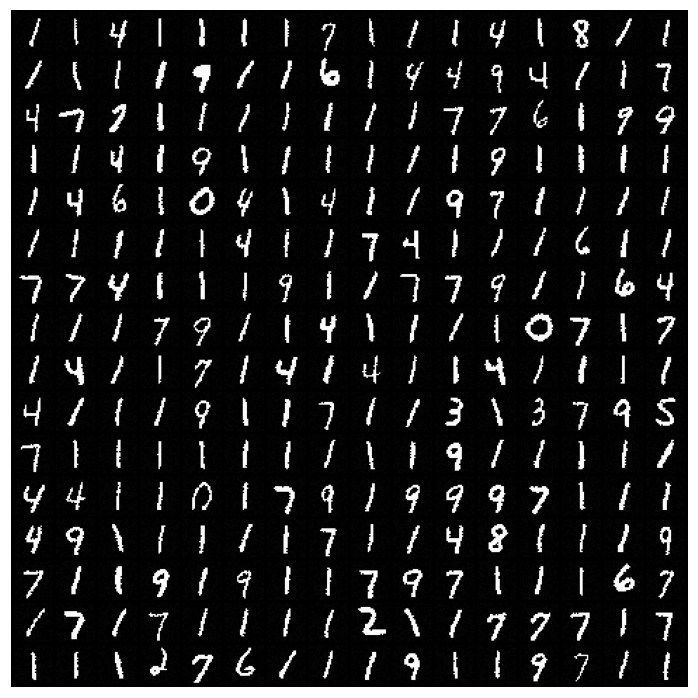}}
\caption{Binarized MNIST samples from Dirichlet diffusion score model (8x time-dilation). Euler Maruyama Sampler with 8x time-dilation and 800 steps. }

\label{fig:mnist_examples8x}
\end{center}
\vskip -0.2in
\end{figure}

\begin{figure}[h!]
\vskip 0.2in
\begin{center}
\centerline{\includegraphics[width=0.5\textwidth]{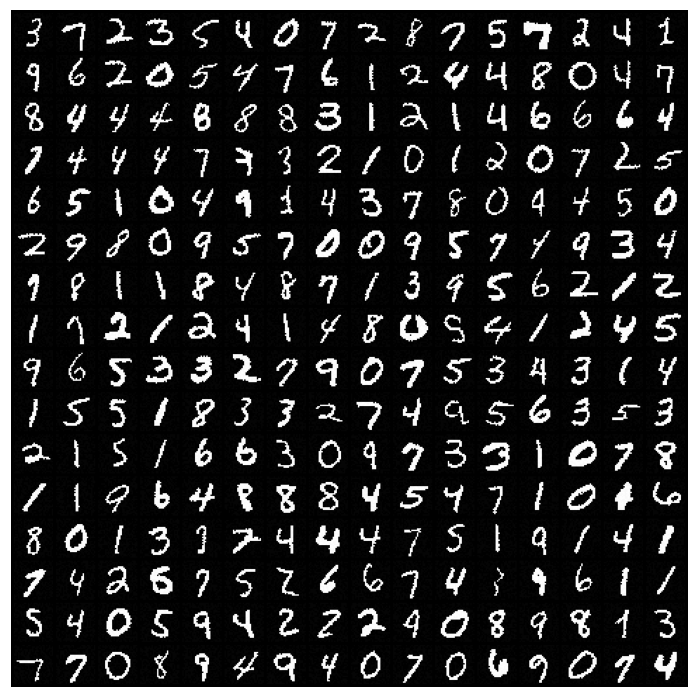}}
\caption{Binarized MNIST samples from Dirichlet diffusion score model (8x time-dilation, time-dilation start time 25\%). Euler Maruyama Sampler with 8x time-dilation started at 25\% time point, with a total of 275 steps. }

\label{fig:mnist_examples8xtd}
\end{center}
\vskip -0.2in
\end{figure}

\pagebreak
\section{Sudoku Transformer Model Performance on the Hard Sudoku Dataset}\label{a:hard_sudoku}

The Sudoku transformer model solved all puzzles in an easy Sudoku dataset with $36$ clues on average \cite{Wang2019-wt} and a hard Sudoku dataset with minimally $17$ clues \cite{Palm2018-recurrent} when multiple samples are used. We also note that the current state-of-the-art results for supervised models are Recurrent Relational Network \cite{Palm2018-recurrent}, which solves $96.7\%$ of the 17-clue Sudoku puzzles within the hard dataset, and SATNET\cite{Wang2019-wt} which solved $98.3\%$ of the easy dataset. Our method solved $100\%$ puzzles, and this requires usually only one or two samples ($\text{mean}=1.19$ samples) on the easy dataset and a high number of samples ($\text{mean}=753$ samples) on the hard dataset. A single sample from our model solves $99.4\%$ of the easy dataset and $42.4\%$ of the hard dataset with 128x time-dilation. Recurrent Relational Network still have better accuracy when only a single sample is allowed for our model. However, our model is never trained on the Sudoku dataset or even in a supervised manner, as it was only trained on generating fully-filled Sudoku from a random Sudoku generator. Thus, these results are not directly comparable. Our result is the first in generating modeling of Sudoku to our knowledge, which already showed very strong performance and even surpass state-of-the-art approaches with supervised training with multiple samples are allowed.

\begin{figure}[h!]\label{fig:hardsudoku}
\vskip 0.in
\begin{center}
\centerline{\includegraphics[width=0.8\columnwidth]{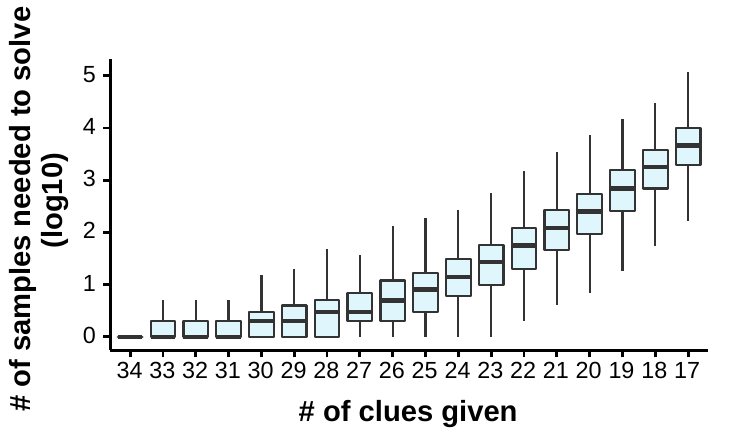}}
\caption{Number of samples required to solve hard Sudoku puzzles by the number of clues. }

\end{center}
\vskip -0.2in
\end{figure}

On the hard dataset, we demonstrated the scaling of the number of samples required with the number of clues given (17 is the minimally possible number for Sudoku) (Figure \ref{fig:hardsudoku}). We note that we are not optimizing this experiment to minimize the number of samples drawn but the overall time spent solving the puzzle, as we can significantly increase single-sample accuracy by applying more time dilation, at the cost of more computation per sample.

This task can serve well as a benchmark for strongly constrained discrete data generation or efficient reverse diffusion sampling methods. With our current setup, each sample is generated with $8$x time-dilation and $200$ steps, which is certainly not optimized given our use of a simple Euler Maruyama sampler modified to support time-dilation.

\pagebreak
\section{Supplementary Information for Promoter Sequence Design}\label{a:promoter_seq_time_dilated}

\subsection{An introduction of the promoter sequence design problem and our study design without assuming prior knowledge of biology}\label{a:intro}

 DNA sequence is composed of 4 bases, or 4 nucleotides: A, C, G, T. The length unit of DNA sequence is basepair (bp) because DNA is usually double-stranded, and each base is paired with its complementary base on the other strand. The DNA base pairing rules are that 1) A and T are complementary to each other, 2) G and C are complementary, 3)the two strands go in opposite directions. Therefore, if a 10bp sequence reads CCAATTTAAG, and the other strand (its reverse complement) will read CTTAAATTGG following the base-pairing rule.

An important function of DNA sequence is to encode genes. For genes to function they have to be first transcribed to RNA (the DNA information is copied basepair-by-basepair to RNA molecules; while a cell has only two copies of the genome DNA, DNA can be transcribed to many RNA molecules). Promoters are sequences that determine where the transcription happens and partially determine how much transcription happens. The amount of transcription from a promoter can be called the \say{expression level} of a promoter. The starting point of transcription, or where DNA starts to be transcribed to RNA, is called the transcription initiation site or the transcription start site. Transcription can happen in many different positions within a promoter, but there is a single basepair that is annotated as the transcription start site (TSS) for every promoter in previously published annotations, which is usually but not always near where most of the transcription starts at. 

Where and how much transcription happens can be measured by experimental methods such as CAGE (Cap analysis gene expression). From experimentally generated data we can obtain a transcriptional initiation signal profile, or more specifically how much transcription happens at every single basepair position, for every human gene promoter. A transcriptional initiation signal profile characterizes the transcriptional function of a promoter, including its expression level which can be obtained from the total amount of transcription over the entire promoter region.

If we have the ability to design promoter sequences that can produce any desired transcription initiation profile, we will also be able to finely control the expression level of any synthetic gene, such as genes producing bioengineering products like antibodies. We can also control very finely where the transcription starts, which may enable potential future applications. On the other hand, it can be a tool for improving our understanding of the mechanism of transcription initiation, or precisely how sequence drives transcription, which is not fully understood.

Therefore, the goal of the promoter sequence design task is to conditionally generate sequences that will produce the same transcription initiation signal profile as the conditional input. To our knowledge, no prior method exists for this task.

To evaluate how well the model works, we first compared the model-designed sequences with real human genome sequences behind transcription initiation signal profiles from the test set. We showed that they indeed have very similar properties, including the base (or nucleotide) composition at every position relative to the transcription start site, as well as the location-specific distribution of known promoter motifs. Known promoter motifs are sequences that are known to frequently appear at promoters and likely play a role in transcription initiation. The examples of the most common promoter motifs are TATA-box (TATAAAA), GC-box (GGGCGGG or CCCGCCC), CCAAT-box (CCAAT or ATTGG). Therefore, we showed that they also show up in our designed sequences and at similar locations with similar frequencies compared to the real genome sequences. We note these are not the only important sequences and our knowledge of sequence rules that drive transcription is still far from complete.

Finally, we used a deep learning model, Sei, that can predict promoter activity to evaluate whether our design sequences will produce the desired activity levels as given in the conditional input: the transcription initiation signal profile. We grouped the transcription initiation signal profiles into 10 groups based on the percentile of expression levels, from the lowest group (0-10\%) to the highest group (90\%-100\%). Encouragingly, the predicted promoter activities of model-designed sequences closely resemble those of the human genome sequence corresponding to each expression group. This suggests that the model can generate both low-activity promoters and high-activity promoters by controlling the input transcription initiation signal profile.

\begin{figure*}[h!]
\vskip 0.2in
\begin{center}
\centerline{\includegraphics[width=\textwidth]{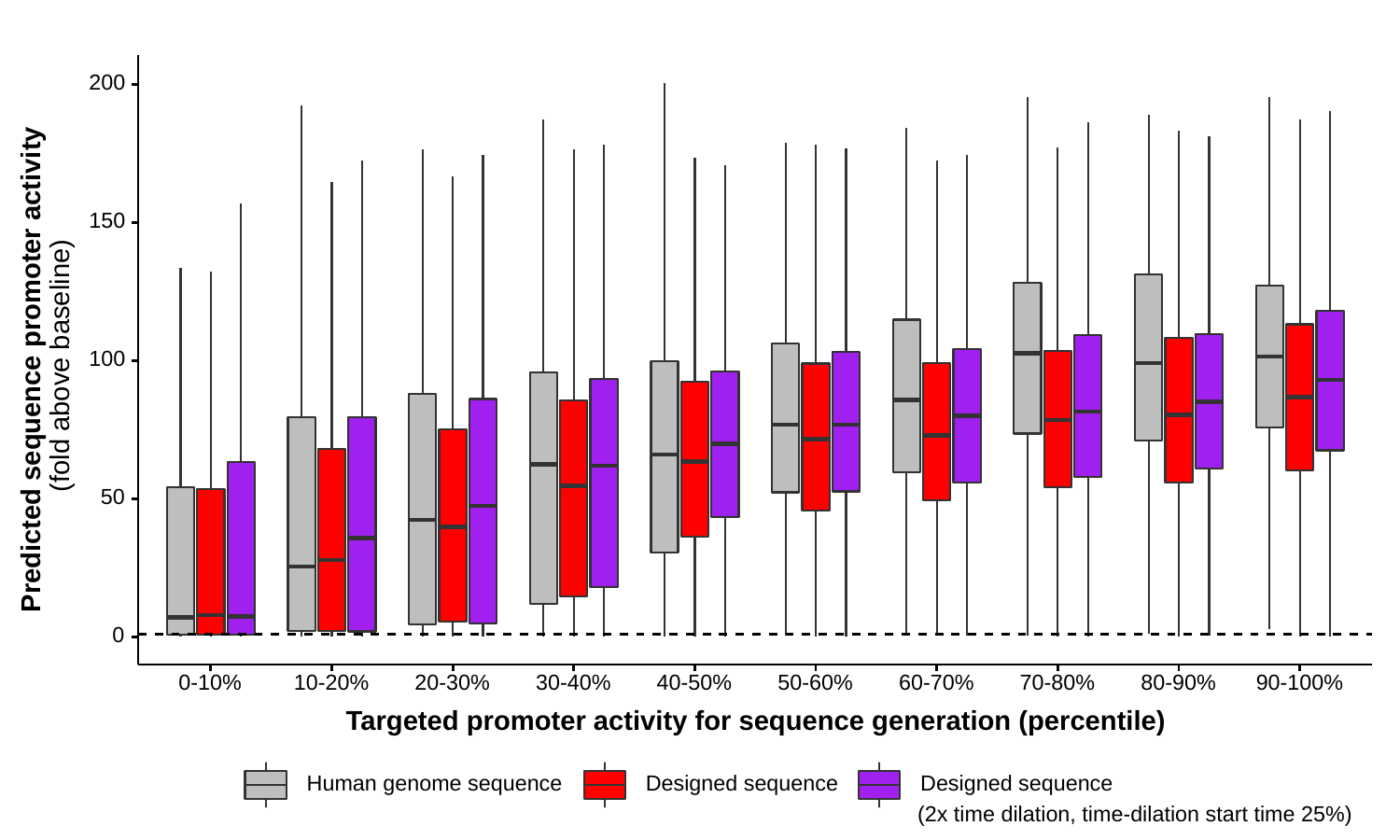}}
\caption{Time-dilation (2x) slightly improves promoter activity predicted from sequence by Sei. Generated sequences are grouped by the targeted promoter activity level (x-axis). Y-axis shows predicted H3K4me4 probability (average across cell types), divided by baseline prediction for average genomic sequences. }
\end{center}
\vskip -0.2in
\end{figure*}

\begin{figure*}[h!]
\vskip 0.2in
\begin{center}
\centerline{\includegraphics[width=\textwidth]{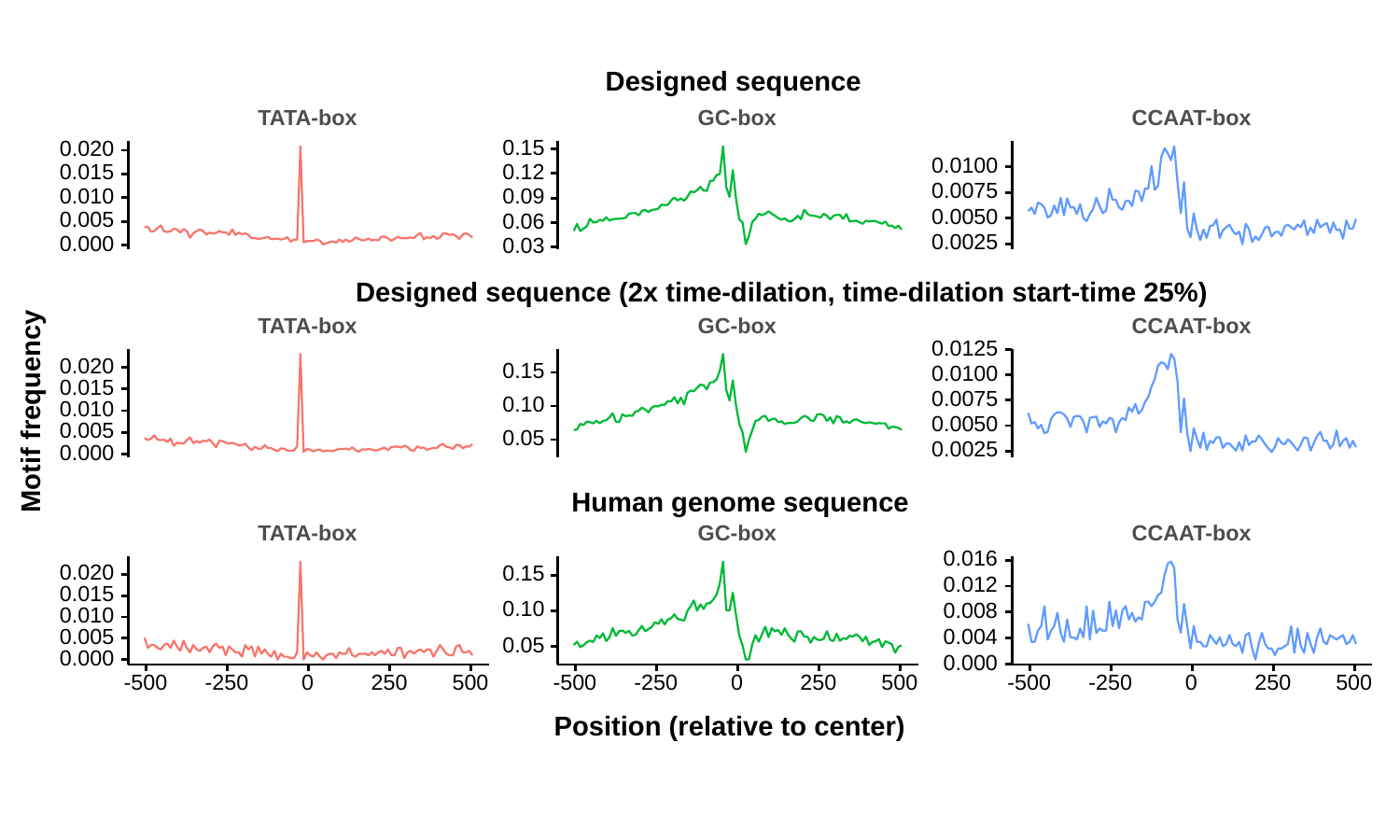}}
\caption{Position-specific motif distribution of designed and generated sequences. The time-dilation increased the known motif frequencies. }

\end{center}
\vskip -0.2in
\end{figure*}

\begin{figure*}[h!]
\vskip 0.2in
\begin{center}
\centerline{\includegraphics[width=\textwidth]{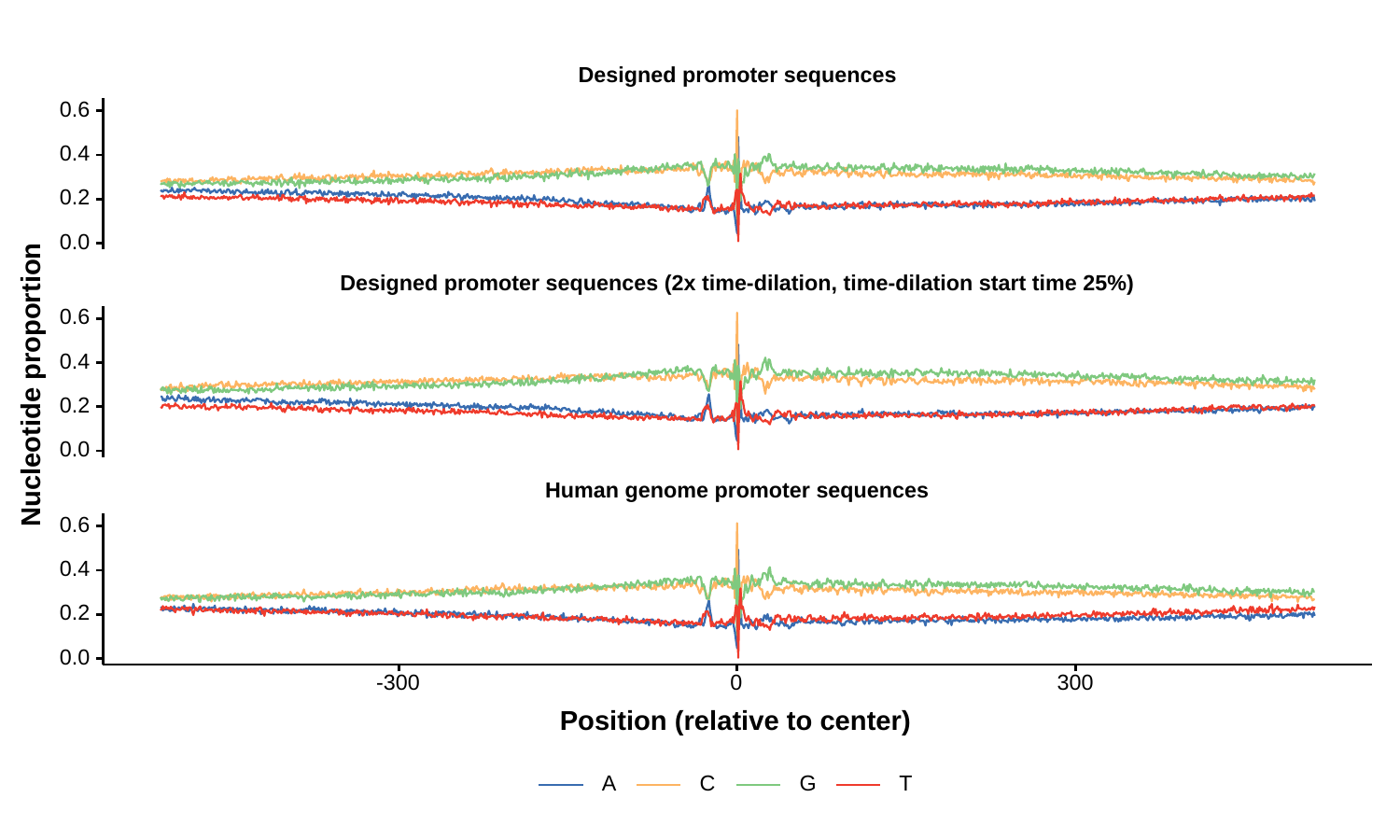}}
\caption{Position-specific nucleotide composition of designed and generated sequences. The time-dilation leads to slightly more biased nucleotide composition. }

\end{center}
\vskip -0.2in
\end{figure*}


\end{document}